\documentclass[11pt]{article}

\usepackage[preprint]{acl}

\usepackage{times}
\usepackage{latexsym}

\usepackage[T1]{fontenc}

\usepackage[utf8]{inputenc}

\usepackage{microtype}

\usepackage{inconsolata}

\usepackage{graphicx}

\usepackage{colortbl}
\usepackage{xcolor}
\usepackage{booktabs}       
\usepackage{amsfonts}       
\usepackage{nicefrac}       
\usepackage{adjustbox}
\usepackage{multirow}
\usepackage{amsmath}
\usepackage{array}
\usepackage{arydshln}
\usepackage{makecell}
\newcommand{\pairdash}{%
  \noalign{%
    \hbox to 0pt{%
      \hspace*{0pt}%
      \vrule width \familycolwidth height 0.4pt depth 0.4pt%
      \kern0pt}%
  }%
  \cmidrule[0.4pt]{3-15}%
}

\usepackage{pifont}
\usepackage[most]{tcolorbox}
\usepackage{tikz}
\usetikzlibrary{positioning,calc,fit,arrows.meta,backgrounds,patterns}
\usepackage{tikz}

\definecolor{bestcell}{HTML}{D4EDDA}      
\definecolor{secondcell}{HTML}{FFF3CD}    
\newcommand{\best}[1]{\cellcolor{bestcell}\textbf{#1}}
\newcommand{\second}[1]{\cellcolor{secondcell}\underline{#1}}

\newcommand{\finding}[2]{%
    \begin{tcolorbox}[
        enhanced,
        colback=white,
        colframe=teal!60!black,
        arc=2pt,
        boxsep=2pt,
        left=6pt, right=6pt, top=3pt, bottom=3pt,
        boxrule=0.8pt,
        drop fuzzy shadow=black!25,
        before skip=4pt, after skip=4pt,
        breakable
    ]
    \noindent\textbf{\textit{\textcolor{teal!60!black}{Finding #1.}}} #2
    \end{tcolorbox}%
}

\definecolor{checkgreen}{HTML}{2E7D4F}
\definecolor{crossred}{HTML}{C44536}
\usepackage{colortbl}
\usepackage[dvipsnames]{xcolor}
\usepackage{amssymb}
\newcommand{\cmark}{\textcolor{checkgreen}{\ding{51}}}
\newcommand{\xmark}{\textcolor{crossred}{\ding{55}}}

\usepackage{xcolor}

\title{How and What to Imagine? Visual Thinking in Unified Multimodal Models for Cross-View Spatial Reasoning}

\author{
   Qian Yang$^{1,2}$, Ankur Sikarwar$^{\ast1,2}$, Huy Le\thanks{Equal Contribution.}$^{1,2}$, Le Zhang$^{1,2}$, \\ \textbf{Zhuan Shi}$^{1,3}$, \textbf{Perouz Taslakian}$^{1,3,4}$,  \textbf{Aishwarya Agrawal}$^{1,2,5}$
    \\
  $^1$ Mila - Québec AI Institute
     $^2$ Université de Montréal    $^3$ McGill University \\ $^4$ ServiceNow AI Research
   $^5$ Canada CIFAR AI Chair 
   \\
    \normalsize {\tt\small \{qian.yang,  aishwarya.agrawal\}@mila.quebec}
}

\begin{document}
\maketitle
\begin{abstract}
Cross-view spatial reasoning remains a weak spot for vision-language models (VLMs): they reason in language and discard the fine-grained geometry the task requires.
Thinking with images aims to fix this by generating an intermediate \emph{thinking-image}, but recent work shows the visual evidence in these traces is largely ignored.
We therefore ask \textbf{how} to make visual thinking matter, and \textbf{what kind} of visual thinking works best. We ask these questions for unified multimodal models (UMMs) that natively support interleaved image–text generation.
For the \textbf{how}, we propose \textbf{View Dropout (VDrop)}, a training-time intervention that hides parts of one input view from the answer span while leaving it visible to the thinking-image tokens.
This incentivizes the model to make use of the thinking-image when answering, rather than answering based on the input views only.
With the thinking-image now being used in answer prediction, we ask \textbf{which kind} of visual thinking works best. We frame this as a \textbf{Learnability–Informativeness (L–I) tradeoff} and 
compare three thinking-image variants: top-down, panoramic, and point-matching renderings.
Trained on synthetic scenes and evaluated on five real-world out-of-domain  benchmarks, panoramic visual thinking with VDrop is the only configuration that is simultaneously informative and learnable, and achieves the best out-of-domain generalization.{\footnote{Code is available at \url{https://github.com/MyLittleChange/VDrop}.}}
\end{abstract}

\section{Introduction}

\emph{Cross-view spatial reasoning} requires inferring scene layout, object placement, and geometry from images taken at different viewpoints. 
It underlies a range of vision-language model (VLM) applications, from embodied agents navigating a room~\cite{wang2025crossviewpointcorrespondencevisionlanguage,DBLP:conf/iccv/HanMHCS25} to video VLMs integrating temporally distant frames~\cite{wu2025spatialmllmboostingmllmcapabilities}, all of which reduce to the same problem: maintaining a consistent scene representation across viewpoints that share only partial visual content. 
We study this capability in its basic form: given two partially overlapping views and a question, a VLM must reason across views to answer correctly, the format adopted by recent multi-view benchmarks~\cite{yang2026mmsibench, jia2026omnispatial, wang2026mindcube}.
Despite strong single-image performance, the strongest VLMs perform marginally above chance at cross-view spatial reasoning~\cite{yang2026mmsibench, jia2026omnispatial}. 
We argue this stems from a representational mismatch: cross-view reasoning is inherently \emph{visual}, yet VLMs reason only through language, verbalizing observations into intermediates that discard the fine-grained geometry the task demands.
Humans, in contrast, reason spatially in the visual domain by mentally constructing internal layouts~\cite{tversky2003structures,levinson2003space, garrod1987saying}, suggesting that letting models \emph{think visually}, by using visual intermediates as part of the reasoning chain, is key to closing this gap.
Existing think-with-image approaches realize this by generating or invoking intermediate visual representations, such as 3D reconstructions, depth maps, or predicted camera trajectories~\cite{yang2026mindjourney, zhang2026think3d, chen2025think}. 
Yet, recent work questions whether these intermediates do real perceptual work: under controlled interventions, model predictions barely change when the visual content of the intermediate is altered, indicating that visual evidence is largely ignored~\cite{liu2025faithfulness}.

\begin{figure*}[th]
\centering
\includegraphics[width=0.98\linewidth]{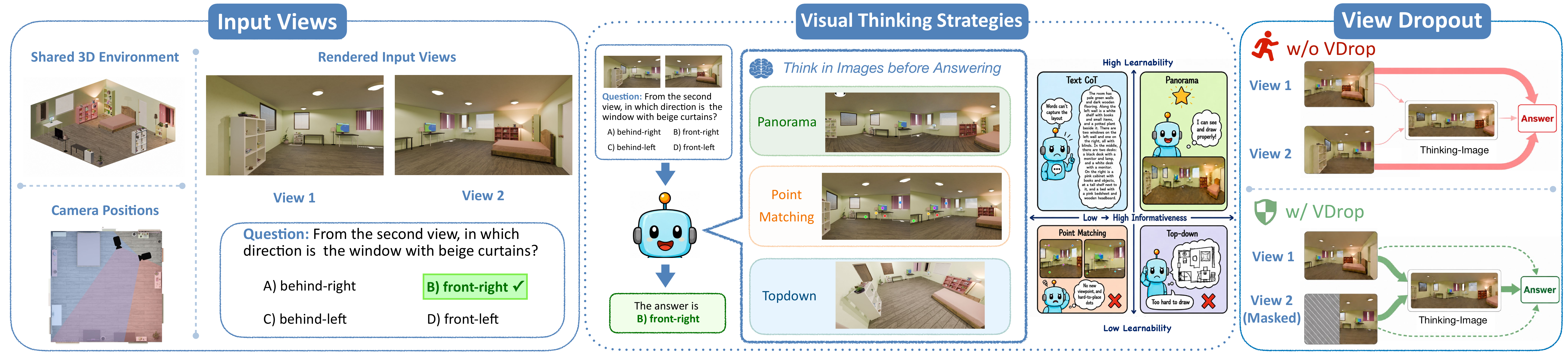}
\caption{\textbf{Visual thinking for cross-view spatial reasoning.} Given two input views and a cross-view spatial question (\textbf{left}), a UMM can generate one of three intermediate thinking-image types (\textbf{middle}) before answering: panorama, point matching, or top-down. \textbf{Right:} without View Dropout, the answer pathway takes a shortcut through the input views, leaving the generated thinking-image unused; with View Dropout, part of one input view is masked, forcing the answer to route through the thinking-image and making visual thinking causally load-bearing.}
\label{fig:vt_strategies}
\label{fig:vdrop}
\vspace{-4mm}
\end{figure*}
Our experiments corroborate that visual evidence is under-used: with standard supervised fine-tuning (SFT), dropping the generated thinking-image at inference barely changes accuracy (Figure~\ref{fig:generate_then_blind}, ``Visual Thinking w/o VDrop''). SFT teaches the model to \emph{generate} a plausible thinking-image but not to \emph{use} it when answering: the thinking-image becomes a decorative by-product of training, present in form but not in function.
This under-use motivates our first research question: (1) how to make visual thinking matter during learning. Once the thinking-image is genuinely used, a second question arises: (2) which kind of thinking-image is most effective for cross-view spatial reasoning, among natural candidates such as panoramic views, top-down layouts, and point-matching overlays that explicitly connect the two views.
To study these questions, we use unified multimodal models (UMMs), which natively generate the thinking-image, enabling end-to-end learning of visual thinking and controlled comparison across intermediate representations within a single model.

To make visual thinking matter during learning, we propose \textbf{View Dropout (VDrop)} (Figure~\ref{fig:vdrop} Right), a training-time intervention that masks part of one input view from the answer span, so the only remaining path for that spatial evidence runs through the generated thinking-image. VDrop requires no architectural change and is agnostic to which thinking-image is generated. Across every thinking-image variant we test, it consistently improves cross-view spatial reasoning.
With VDrop in place, candidate thinking-image representations become meaningfully comparable, and we ask which works best. 
The different thinking-image variants trade off along two axes that prior work has not cleanly separated: \textbf{informativeness} (how much spatial structure the thinking-image variant unveils) and \textbf{learnability} (how reliably the UMM can produce that variant). 
We formalize this as a Learnability–Informativeness (\textit{L–I}) tradeoff: a thinking-image type benefits spatial reasoning only if it is both spatially informative \textit{and} learnable from data alone, and neither axis is sufficient on its own. 
We instantiate this study with synthetic scenes from Infinigen Indoors, {testing on BAGEL \cite{deng2025bagel} and SenseNova-U1 \cite{sensenova2026sensenovau1}}, and evaluating on one in-domain synthetic benchmark and five real-world out-of-domain benchmarks.
Experiments show that visual thinking improves cross-view spatial reasoning both in- and out-of-domain, and that VDrop makes the generated thinking-image causally load-bearing, consistently improving OOD performance {on both UMMs}.
Trained on only 8K synthetic samples, our best configuration achieves a $6.7$-point OOD gain over vanilla BAGEL and surpasses all prior methods we compare against, including methods trained on at least $3\times$ more data. 
Once visual thinking is forced to matter, the choice of representation also matters: top-down views, though informative, are not directly learnable by current UMMs, leaving panoramic visual thinking as the only candidate that scores high on both \textit{L–I} axes and the only one that consistently beats prior methods on OOD generalization.

Our contributions are as follows:


\noindent\textbullet~\textbf{Method.} We identify under-use as a pervasive failure mode of visual thinking and propose \textbf{View Dropout}, a training-time intervention that requires no architectural change, is agnostic to the thinking-image type, and consistently improves OOD cross-view spatial reasoning across all three thinking-image variants {and across two architecturally distinct UMMs}.

\noindent\textbullet~\textbf{Framework.} We frame the choice of visual thinking as a \emph{Learnability–Informativeness tradeoff}, disentangling two axes that prior work has conflated: a representation may fail either because it does not encode sufficient information or because it is difficult to learn.

\noindent\textbullet~\textbf{Empirical analysis.} On one in-domain synthetic benchmark and five real-world OOD benchmarks, we show that the thinking-image becomes causally used only after VDrop training, and that panoramic visual thinking is the most informative and learnable representation. With only 8K training samples, it outperforms prior BAGEL-based visual-thinking methods trained on at least $3\times$ more data.
\section{Related Work}
\vspace{-2mm}

\noindent\textbf{Cross-View Spatial Reasoning.}
Cross-view spatial reasoning, the task of reasoning about object positions, distances, depth ordering, and viewpoint relationships across multiple views, has emerged as a documented weakness of current VLMs~\cite{yang2026mmsibench, jia2026omnispatial,li2025unfoldingspatialcognitionevaluating, fu2024blink, zhang2026things}, with even strong open-source VLMs scoring only marginally above chance.
Existing remedies primarily target the language pathway: spatial instruction tuning~\cite{chen2024spatialvlm, sensenova-si} and curated reasoning-trace fine-tuning train VLMs to verbalise spatial structure into text. Other approaches add architectural components, injecting spatial priors via depth or 3D-aware encoders~\cite{thai2025splattalk}, but still rely on language for the reasoning itself.
Across both lines, the reasoning remains \emph{linguistic}: the scene is verbalised into text, discarding the fine-grained geometry the task requires.
Our work investigates a different axis, whether models can be trained to reason \emph{visually} by generating intermediate visual representations of the scene.

\noindent\textbf{Think with Images.}
Recent works show that models can ``think in images'' by sketching annotations on inputs or composing reasoning chains from generated images~\cite{hu2024visual, cheng2026visual, xu2026visual}.
A parallel line introduces 3D-derived intermediates, such as depth maps, Gaussian splats, or 3D reconstructions~\cite{zhang2026think3d, chen2025think}, while others predict camera trajectories to mentally simulate unseen viewpoints~\cite{yang2026mindjourney, yu2026and}.
However, the visual content in these intermediates is often ignored by the answer pathway~\cite{liu2025faithfulness}, questioning whether the generated image is doing real perceptual work.
Our work takes this critique as a starting point and asks two questions prior work leaves open: \emph{how} to train models so that the thinking-image is causally used, and \emph{which kind} of thinking-image is most effective once it is.
{Our fix can be seen as shortcut removal~\cite{geirhos2020shortcut}: VDrop blocks the input-view shortcut during training, while keeping the masked evidence reachable through the model's own generated thinking-image. Closest to ours, Markovian training forces the answer to route through a textual CoT~\cite{viteri2026markovian}; VDrop enforces the same dependence on a visual intermediate.}

\noindent\textbf{Unified Multimodal Models.}
Unified multimodal models (UMMs) extend the single-encoder/decoder paradigm of standard VLMs to support \emph{interleaved} image--text generation within one architecture~\cite{xie2025show, wu2025janus, chen2025janus, deng2025bagel, liu2025tunatamingunifiedvisual, tuna2, sensenova2026sensenovau1}.
Designs differ in how they reconcile the conflicting representational needs of understanding and generation: Janus~\cite{wu2025janus, chen2025janus} decouples visual encoding into two specialised pathways feeding a shared transformer; 
BAGEL~\cite{deng2025bagel} couples a multimodal understanding encoder with a diffusion-based image generator via a unified token interface; 
{SenseNova-U1~\cite{sensenova2026sensenovau1} instead removes pretrained visual encoders altogether, consuming raw pixels directly and generating images via pixel flow matching rather than a VAE-based diffusion head}.
Recent work already uses UMMs as backbones for interleaved visual reasoning, training them to generate intermediate images during decoding~\cite{gu2025thinkmorph, li2026zebracot}. This native generation capability makes UMMs a natural testbed for our study: a single model can produce and reason over thinking-images end-to-end, enabling controlled comparison across thinking-image types without external tools.
Following ThinkMorph~\cite{gu2025thinkmorph}, we conduct our experiments on BAGEL {and further validate cross-architecture transfer on SenseNova-U1}.
\vspace{-2mm}
\section{Method}
\label{sec:framework}
\vspace{-2mm}

\subsection{Unified Multimodal Models}
\label{sec:umm}
Given two input views $V_1, V_2$ and a textual question $q$, a UMM generates an output sequence $\mathbf{o} = (o_1, o_2, \ldots, o_T)$ where each $o_t$ is either a text token or an image token drawn from a shared vocabulary space.
This allows the model to produce an intermediate visual representation $I_{\text{vt}}$ (the \emph{thinking-image}; the subscript denotes ``visual thinking'') as part of its reasoning before producing the final textual answer $a$.
We refer to the full process as \emph{visual thinking}, and to a single sequence $(V_1, V_2, q) \rightarrow I_{\text{vt}} \rightarrow a$ as a \emph{visual-thinking trace}.
Given a dataset of such traces $\mathcal{D} = \{(V_1^{(i)}, V_2^{(i)}, q^{(i)}, I_{\text{vt}}^{(i)}, a^{(i)})\}_{i=1}^{N}$, supervised fine-tuning trains the UMM to generate the interleaved sequence end-to-end: first the thinking-image $I_{\text{vt}}$ conditioned on the inputs, then the answer $a$ conditioned on the inputs and the generated thinking-image.

\subsection{View Dropout}
\label{sec:vdrop}
\noindent\textbf{Motivation.}
Standard SFT supervises both the thinking-image and the answer, but does not enforce that the answer tokens \emph{depend} on $I_{\text{vt}}$ when reasoning. Thus, the model can successfully minimise the thinking-image generation loss and the answer loss without actually making use of the thinking-image while answering,
leaving the generated thinking-image as a decorative side-product.
Recent analyses~\cite{liu2025faithfulness} report that predictions remain nearly unchanged under visual intervention, indicating that the visual evidence in the thinking-image is largely ignored.

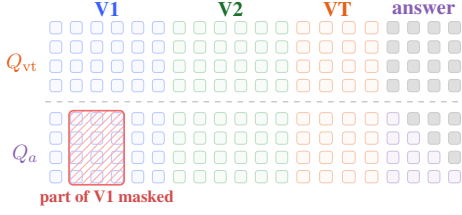
\begin{figure}[t]
\centering
\resizebox{\linewidth}{!}{%
\begin{tikzpicture}[
    x=0.85cm,
    y=0.72cm,
    >=Latex,
    font=\sffamily\large
]

\definecolor{myblue}{RGB}{55,95,255}
\definecolor{mygreen}{RGB}{33,140,55}
\definecolor{myorange}{RGB}{230,95,20}
\definecolor{maskred}{RGB}{235,85,85}
\definecolor{myviolet}{RGB}{135,95,180}
\definecolor{myteal}{RGB}{20,125,140}

\node[text=myblue,font=\bfseries\normalsize]           at (14.555,7.40) {V1};
\node[text=mygreen!80!black,font=\bfseries\normalsize] at (17.265,7.40) {V2};
\node[text=myteal,font=\bfseries\normalsize]           at (19.300,7.40) {q};
\node[text=myorange,font=\bfseries\normalsize]         at (20.970,7.40) {VT};
\node[text=myviolet,font=\bfseries\normalsize]         at (22.865,7.40) {answer};

\node[text=myorange,font=\bfseries\normalsize,anchor=east]
    at (13.20,6.01) {$Q_{\mathrm{vt}}$};
\node[text=myviolet,font=\bfseries\normalsize,anchor=east]
    at (13.20,3.65) {$Q_{a}$};

\foreach \i in {0,...,5} {
    \foreach \j in {0,...,3} {
        \draw[rounded corners=0.05cm, draw=myblue!35,
              line width=0.5pt, fill=myblue!3]
            (13.30+0.45*\i, 5.26+0.50*\j) rectangle ++(0.26,0.32);
    }
}
\foreach \i in {0,...,5} {
    \foreach \j in {0,...,3} {
        \draw[rounded corners=0.05cm, draw=mygreen!30,
              line width=0.5pt, fill=mygreen!3]
            (16.01+0.45*\i, 5.26+0.50*\j) rectangle ++(0.26,0.32);
    }
}
\foreach \i in {0,...,2} {
    \foreach \j in {0,...,3} {
        \draw[rounded corners=0.05cm, draw=myteal!40,
              line width=0.5pt, fill=myteal!4]
            (18.72+0.45*\i, 5.26+0.50*\j) rectangle ++(0.26,0.32);
    }
}
\foreach \i in {0,...,3} {
    \foreach \j in {0,...,3} {
        \draw[rounded corners=0.05cm, draw=myorange!35,
              line width=0.5pt, fill=myorange!3]
            (20.08+0.50*\i, 5.26+0.50*\j) rectangle ++(0.28,0.32);
    }
}
\foreach \i in {0,...,3} {
    \foreach \j in {0,...,3} {
        \draw[rounded corners=0.05cm, draw=gray!40,
              line width=0.5pt, fill=gray!25]
            (22.06+0.45*\i, 5.26+0.50*\j) rectangle ++(0.26,0.32);
    }
}

\draw[dashed, gray!55, line width=0.6pt]
    (13.20,5.0) -- (23.76,5.0);

\foreach \i in {0,...,5} {
    \foreach \j in {0,...,3} {
        \draw[rounded corners=0.05cm, draw=myblue!35,
              line width=0.5pt, fill=myblue!3]
            (13.30+0.45*\i, 2.90+0.50*\j) rectangle ++(0.26,0.32);
    }
}
\foreach \i in {0,...,5} {
    \foreach \j in {0,...,3} {
        \draw[rounded corners=0.05cm, draw=mygreen!30,
              line width=0.5pt, fill=mygreen!3]
            (16.01+0.45*\i, 2.90+0.50*\j) rectangle ++(0.26,0.32);
    }
}
\foreach \i in {0,...,2} {
    \foreach \j in {0,...,3} {
        \draw[rounded corners=0.05cm, draw=myteal!40,
              line width=0.5pt, fill=myteal!4]
            (18.72+0.45*\i, 2.90+0.50*\j) rectangle ++(0.26,0.32);
    }
}
\foreach \i in {0,...,3} {
    \foreach \j in {0,...,3} {
        \draw[rounded corners=0.05cm, draw=myorange!35,
              line width=0.5pt, fill=myorange!3]
            (20.08+0.50*\i, 2.90+0.50*\j) rectangle ++(0.28,0.32);
    }
}
\foreach \i in {0,...,3} {
    \foreach \j in {0,...,3} {
        \pgfmathtruncatemacro{\sum}{\i+\j}
        \ifnum\sum>3
            \draw[rounded corners=0.05cm, draw=gray!40,
                  line width=0.5pt, fill=gray!25]
                (22.06+0.45*\i, 2.90+0.50*\j) rectangle ++(0.26,0.32);
        \else
            \draw[rounded corners=0.05cm, draw=myviolet!40,
                  line width=0.5pt, fill=myviolet!6]
                (22.06+0.45*\i, 2.90+0.50*\j) rectangle ++(0.26,0.32);
        \fi
    }
}

\filldraw[rounded corners=0.08cm,
          fill=maskred!10, draw=maskred!85, line width=0.9pt,
          pattern=north east lines, pattern color=maskred!45]
    (13.72,2.85) rectangle (14.93,4.77);

\node[text=maskred!90!black,font=\bfseries\small,
      align=center] at (14.55,2.50)
    {part of V1 masked};
\end{tikzpicture}%
}
\caption{\textbf{VDrop attention mask.} Answer queries $Q_a$ cannot attend to the masked region (red hatched), while thinking-image queries $Q_{\mathrm{vt}}$ retain full access to all. }
\label{fig:vdrop_mask}
\vspace{-4mm}
\end{figure}

\noindent\textbf{Method overview.}
To force the thinking-image to be a load-bearing component of reasoning, we introduce \emph{View Dropout (VDrop)}, 
a training-time intervention that hides a randomly selected contiguous region of one input view\footnote{We mask only one of the two views; masking both removes too much spatial evidence and performs worse in our ablations (Appx.~\ref{sec:appendix_vdrop_ablation}).} from the answer tokens, while leaving the thinking-image tokens fully visible (Figure~\ref{fig:vdrop_mask}).
Under standard SFT, the layout information needed to answer is fully available across the two input views, so the model is not compelled to rely on the thinking-image. 
VDrop removes this shortcut: with part of one view hidden from the answer tokens, the complete layout is recoverable only from the thinking-image, so the answer pathway must attend to it.

\noindent\textbf{Attention mask construction.}
Let $M$ denote the standard per-sample attention mask, with $M_{qk} = 0$ permitting attention from query $q$ to key $k$ and $M_{qk} = -\infty$ blocking it, and let $Q_a$ denote the query positions of the answer span.
We sample a primary view $v \in \{V_1, V_2\}$ uniformly and a contiguous subset $D_v$ of its patch-token positions, and edit the mask so the answer cannot read $D_v$: $M_{qk} \leftarrow -\infty$ for all $q \in Q_a$ and $k \in D_v$.
For thinking-image queries, the mask entries to $V_1$ and $V_2$ are unchanged, so $I_{\text{vt}}$ generation continues to attend fully to both input views (see Figure~\ref{fig:vdrop_mask}).

\noindent\textbf{Region selection.}
We sample $D_v$ as a contiguous axis-aligned rectangle of patch positions on the chosen view $v$. 
By hiding a coherent chunk of the scene rather than scattered patches, the answer pathway cannot interpolate the missing region from nearby patches and must recover it from the thinking-image. 
In BAGEL, each input view is encoded into two parallel token streams: ViT tokens, which carry semantic content for understanding, and VAE tokens, which carry pixel-level detail for generation. A masked region must therefore be hidden in \emph{both} streams; masking only one would let the answer recover the region through the other. 
We thus mask the ViT and VAE tokens covering $D_v$ jointly. $D_v$ covers a fixed fraction $\rho$ of patch positions on view~$v$.\footnote{We set $\rho = 0.5$ in all main experiments. We sweep over $\rho$ and compare contiguous-region masking against random-patch masking in Appx.~\ref{sec:appendix_vdrop_ablation}; contiguous-region masking at $\rho = 0.5$ gives the best OOD accuracy.}

\noindent\textbf{Training curriculum.}
Applying the mask from the first step collapses learning: the model is asked to route evidence through $I_{\text{vt}}$ before SFT has shaped what $I_{\text{vt}}$ should encode.
We therefore anneal the masking probability $p_{\mathrm{mask}}(s)$ over training steps $s$: $p_{\mathrm{mask}} = 0$ for $s < s_w$ (warmup), $p_{\mathrm{mask}} = (s - s_w) / s_a$ for $s_w \le s < s_w + s_a$ (linear anneal), and $p_{\mathrm{mask}} = 1$ thereafter.
Warmup lets the model first learn to generate and use $I_{\text{vt}}$ alongside the full input; annealing then introduces VDrop pressure gradually, so a working $I_{\text{vt}} \rightarrow a$ route is in place by the time the input views are fully masked.\footnote{We use $s_{\mathrm{w}} = 500$ and $s_{\mathrm{a}} = 1500$; Appx.~\ref{sec:appendix_vdrop_ablation} ablates these values.}

\noindent\textbf{Compatibility.}
VDrop modifies only the answer-side attention mask and leaves the SFT objective unchanged. The thinking-image is generated from the full input views and supervised toward its ground-truth render, as under standard SFT; VDrop changes only whether the answer is \emph{forced to use} the thinking-image, not how it is generated. It is compatible with any thinking-image strategy.

\vspace{-2mm}
\subsection{Visual Thinking Strategies}
\label{sec:strategies}

To study which type of thinking-image is most effective for cross-view spatial reasoning, we consider three visual-thinking variants, each capturing a distinct strategy for bridging the two input views (Figure~\ref{fig:vt_strategies}).
\textbf{Panoramic view}: a wide-angle rendering from the observer's pose, \emph{reconstructs the full scene} so that $V_1$ and $V_2$ become sub-regions of one unified visual field. 
\textbf{Top-down view}: a high-angle rendering from a top corner of the room, \emph{lifts to a shared external frame} that exposes the global layout while still revealing object sides and depth.
\textbf{Point matching}: the two input views shown side by side with coloured markers on corresponding objects, \emph{makes cross-view correspondences explicit} without changing the camera frame.

\noindent
Together, these variants span a natural design space: panorama unifies the views into one scene, top-down reprojects them into a shared external frame, and point matching annotates the views in place with cross-view identity. 
We deliberately avoid intermediates that require auxiliary modules, such as depth maps or 3D reconstructions, to isolate the contribution of the thinking-image itself rather than the supervision signal of an external tool.

\begin{table}[t]
\centering
\small
\adjustbox{max width=0.85\columnwidth}{%
\begin{tabular}{lp{0.35\linewidth}p{0.4\linewidth}}
\toprule[1.5pt]
\textbf{Type} & \textbf{Task description} & \textbf{Example question} \\
\midrule
Anchor & Identify an object visible in both views. & ``Which object appears in both views?'' \\
Counting & Count total instances of a given object across views. & ``How many chairs are in the scene?'' \\
Relative Distance & Identify the closest or farthest object from a reference. & ``Which object is closest to the desk?'' \\
Relative Direction & Identify the direction of an object relative to a reference. & ``Which side of the sofa is the lamp on?'' \\
\bottomrule[1.5pt]
\end{tabular}
}
\caption{The four cross-view question types in our 8K Infinigen training set.}
\label{tab:question_types}
\vspace{-4mm}
\end{table}

\vspace{-2mm}
\subsection{Training Data: Infinigen Indoors}
\label{sec:data}
To obtain clean training signal for each visual-thinking strategy, we construct training data from Infinigen Indoors~\cite{infinigen}, whose procedural 3D annotations yield unambiguous ground-truth answers and ground-truth thinking-images.
Each scene provides two egocentric views with overlapping fields of view, along with the corresponding top-down, panoramic, and point-matching renderings used as ground-truth $I_{\text{vt}}$.
Following the COSMIC benchmark~\cite{sikarwar2026communicating}, we construct four cross-view question types (Table~\ref{tab:question_types}) that require integrating spatial information from both views.
The full training set contains 7{,}921 QA pairs across 1{,}584 unique scenes; per-type descriptions, scene-generation details, and trace construction are in Appx.~\ref{sec:appendix_training_data}.
\vspace{-2mm}
\section{Experiments}
\label{sec:findings}
\vspace{-2mm}
\subsection{Experimental Setup}
\label{sec:impl}

\noindent\textbf{Model and baselines.}
{Our main experiments fine-tune \emph{BAGEL}~\cite{deng2025bagel}, a SOTA open-source UMM built on a Mixture-of-Transformers architecture ($14$B total parameters, $7$B active per token); for cross-architecture validation (\S\ref{sec:sensenova}) we additionally fine-tune \emph{SenseNova-U1}~\cite{sensenova2026sensenovau1}, an encoder-free $8$B UMM natively trained for interleaved vision--language generation.} We compare four categories of baselines against our visual-thinking variants.
First, \emph{vanilla BAGEL} without fine-tuning, measuring the gain attributable to visual-thinking SFT.
Second, two non-visual-thinking baselines fine-tuned on the same 8K synthetic Infinigen data as our visual-thinking variants: \emph{No-Think}, which answers directly from $V_1, V_2$ and the question, and \emph{Text CoT}, which produces a textual CoT instead of an image, annotated by prompting a strong off-the-shelf VLM with the input views and the ground-truth answer (details in Appx.~\ref{app:text_cot_anno}).
These isolate the contribution of generating an intermediate \emph{image} from fine-tuning alone or a textual intermediate.
Third, two BAGEL-based visual-thinking methods that fine-tune BAGEL without VDrop: \emph{ThinkMorph}~\cite{gu2025thinkmorph}, which continues training BAGEL on $\sim$24K interleaved reasoning traces, and \emph{BAGEL-Zebra-CoT}~\cite{li2026zebracot}, which fine-tunes BAGEL on 182K interleaved text-image reasoning traces.
Fourth, \emph{Qwen3-VL}~\cite{Qwen3-VL}, a strong understanding-only VLM, to contextualise the gap between standard VLMs and visual-thinking-trained UMMs.
All baselines use the same multiple-choice prompt and answer-extraction protocol.

\noindent\textbf{Training hyperparameters.}
We apply LoRA fine-tuning on BAGEL with rank $32$ and alpha $64$, training for $7{,}000$ steps on $4\times$H100 GPUs. We use the Adam optimizer with a learning rate of $1\times10^{-5}$ and a cosine-decay schedule, and weight the cross-entropy and MSE losses equally ($1.0$ each). The maximum context length is $35{,}000$ tokens for text-only training (No-Think and Text CoT) and $20{,}000$ tokens for visual-thinking training (panoramic, top-down, and point-matching, each with and without VDrop).
{For SenseNova-U1, whose framework lacks LoRA support, we instead unfreeze the top six LLM layers, training for $490$ steps with learning rate $5\times10^{-5}$ and a $14{,}336$-token context.}

\definecolor{avgcol}{HTML}{E8F0F5}

\definecolor{bestbadge}{HTML}{D4EDDA}
\definecolor{secondbadge}{HTML}{FFF3CD}
\definecolor{besttext}{HTML}{1B7F3A}
\definecolor{secondtext}{HTML}{8A6D00}
\definecolor{oursheader}{HTML}{ECEFF4}

\newlength{\familycolwidth}
\setlength{\familycolwidth}{2.95cm}
\definecolor{famA}{HTML}{FFFFFF}
\definecolor{famB}{HTML}{FFFFFF}
\definecolor{famC}{HTML}{FFFFFF}
\definecolor{famD}{HTML}{FFFFFF}
\newcommand{\famblank}[1]{\cellcolor{#1}}

\newcommand{\famlabel}[3]{%
  \cellcolor{#2}%
  \multirow{-#1}{\familycolwidth}{%
    \centering\bfseries\makecell[c]{#3}%
  }%
}
\newcommand{\rankbadge}[2]{%
  \begingroup
  \setlength{\fboxsep}{1.5pt}%
  \colorbox{#1}{\strut #2}%
  \endgroup
}

\renewcommand{\best}[1]{%
  \rankbadge{bestbadge}{\textbf{\textcolor{besttext}{#1}}}%
}

\renewcommand{\second}[1]{%
  \rankbadge{secondbadge}
  {\underline{\textcolor{secondtext}{#1}}}%
}

\begin{table*}[t]
\centering
\adjustbox{max width=\textwidth}{%
\begin{tabular}{>{\centering\arraybackslash}m{\familycolwidth}lcccccc|ccccccc}
\toprule[1.5pt]
 & \textbf{Model} & \textbf{VDrop}
& \multicolumn{4}{c}{\textbf{COSMIC (in-domain)}}
& \cellcolor{avgcol}\textbf{Avg.}
& \textbf{MMSI}
& \textbf{MindCube}
& \multicolumn{2}{c}{\textbf{OmniSpatial}}
& \textbf{STARE}
& \textbf{BLINK}
& \cellcolor{avgcol}\textbf{Avg.} \\
\cmidrule(lr){4-7} \cmidrule(lr){9-14}
& 
& 
& \textbf{Anchor} & \textbf{Count.} & \textbf{Rel-Dist.} & \textbf{Rel-Dir.}
& \cellcolor{avgcol}\textbf{ID}
& \textbf{Overall} & \textbf{Overall}
& \textbf{CL} & \textbf{PT}
& \textbf{Persp.}
& \textbf{MultiView}
& \cellcolor{avgcol}\textbf{OOD} \\
\midrule

\famblank{famA}
&Qwen3-VL-4B & \xmark
& 62.8 & 44.4 & 40.4 & 22.4 & \cellcolor{avgcol}42.5
& 27.4 & 29.0 & 28.6 & 40.1 & 29.2 & 48.9 & \cellcolor{avgcol}33.8 \\

\famlabel{2}{famA}{Understanding\\VLMs}
&Qwen3-VL-8B & \xmark
& 64.8 & 54.4 & 40.8 & 26.8 & \cellcolor{avgcol}46.7
& 28.0 & 34.4 & 25.4 & \best{45.5} & 31.6 & 55.6 & \cellcolor{avgcol}37.0 \\

\midrule
\famblank{famB}
&BAGEL & \xmark
& 18.0 & 42.1 & 24.4 & 21.2 & \cellcolor{avgcol}26.4
& 26.9 & 31.7 & 31.1 & 38.9 & 28.0 & 45.1 & \cellcolor{avgcol}33.3 \\
\famblank{famB}
&BAGEL-Zebra-CoT (182K)& \xmark
& 8.8 & 24.4 & 28.4 & 24.8 & \cellcolor{avgcol}21.6
& 23.2 & 21.7 & 29.0 & 43.0 & 28.4 & 24.8 & \cellcolor{avgcol}26.8 \\
\famlabel{3}{famB}{BAGEL-based \\ prior work }
&ThinkMorph (24K) & \xmark
& 49.6 & 43.6 & 32.0 & 30.0 & \cellcolor{avgcol}38.8
& 26.5 & \second{39.2} & 33.7 & 44.6 & 28.8 & 52.6 & \cellcolor{avgcol}37.2 \\

\midrule
\famblank{famC}
&No-Think & \xmark
& 86.8 & 82.4 & 67.6 & 85.6 & \cellcolor{avgcol}80.6
& 27.4 & \best{41.1} & 30.2 & 44.0 & 24.8 & 45.1 & \cellcolor{avgcol}35.1 \\
\famlabel{2}{famC}{BAGEL w/ 8K\\non-visual}
&Text CoT & \xmark
& 56.0 & 60.4 & 44.4 & 40.8 & \cellcolor{avgcol}50.4
& 24.6 & 25.3 & 29.4 & 35.8 & 26.0 & 54.9 & \cellcolor{avgcol}32.7 \\

\midrule
\famblank{famD}
&\multirow{2}{*}{Panoramic}
& \xmark
& \second{93.6} & \best{83.6} & \second{76.4} & 87.2 & \cellcolor{avgcol}{85.2}
& 24.9 & 36.9 & 32.9 & 43.9 & 32.4 & 55.6 & \cellcolor{avgcol}\second{37.6} \\

\famblank{famD}
& & \cmark
& 89.2 & 78.8 & 74.8 & \second{93.2} & \cellcolor{avgcol}84.0
& 26.0 & 34.1 & \best{38.9} & \second{45.3} & \best{35.6} & \best{62.4} & \cellcolor{avgcol}\best{40.0}\,\textsuperscript{\textcolor{teal}{(+2.4)}} \\

\cdashline{2-15}
\famblank{famD}
&\multirow{2}{*}{Point Matching}
& \xmark
& 92.8 & 81.2 & 73.2 & \best{94.4} & \cellcolor{avgcol}\best{85.4}
& 27.8 & 35.2 & 31.0 & 41.7 & 24.0 & 52.6 & \cellcolor{avgcol}35.2 \\

\famblank{famD}
& & \cmark
& 92.8 & 78.8 & \best{78.0} & 91.6 & \cellcolor{avgcol}\second{85.3}
& 28.3 & 34.1 & \second{33.7} & 43.3 & \second{34.4} & 45.1 & \cellcolor{avgcol}36.1\,\textsuperscript{\textcolor{teal}{(+0.9)}} \\

\cdashline{2-15}
\famblank{famD}
& \multirow{2}{*}{Top-down}
& \xmark
&93.6  &\second{83.2}  & 68.4    & 92.0 &\cellcolor{avgcol}84.3
&\second{28.8}      & 35.2          & 31.4    & 39.8   & 28.0 & \second{58.7} & \cellcolor{avgcol}37.3 \\

\famlabel{6}{famD}{BAGEL w/ 8K \\visual thinking}
& & \cmark
&\best{94.4}    & 76.8  & 72.0     & 89.2 &\cellcolor{avgcol}83.1
&\best{32.0} & 36.5    & 32.5    & 44.2   & 26.0 & 57.1  &\cellcolor{avgcol}38.0\,\textsuperscript{\textcolor{teal}{(+0.7)}}
\\

\bottomrule[1.5pt]
\end{tabular}%
}
\caption{\textbf{Comparison with baselines and VDrop ablation across thinking-image types.}
The two blue \textbf{Avg.}\ columns report the in-domain mean over the four COSMIC subtasks and the out-of-domain mean over five real-world benchmarks.
Per column, the \best{best} and \second{second-best} accuracy are highlighted.
}
\label{tab:vdrop_with_without}
\vspace{-5mm}
\end{table*}

\noindent\textbf{VDrop hyperparameters.}
View Dropout is controlled by three hyperparameters: the warmup length $s_{\mathrm{w}}$, the anneal length $s_{\mathrm{a}}$, and the drop fraction $\rho = |D_v| / |V_v|$ that determines the proportion of the chosen view's patch tokens hidden from the answer span.
Unless stated otherwise, we use $s_{\mathrm{w}} = 500$, $s_{\mathrm{a}} = 1500$, and $\rho = 0.5$, with the \emph{contiguous region} masking strategy as the default selection rule.
The primary view $v$ is sampled uniformly from $V_1$ and $V_2$ at each training step.
{On SenseNova-U1, we keep the same masking strategy and $\rho$, and scale the curriculum to its shorter run ($s_{\mathrm{w}} = 50$, $s_{\mathrm{a}} = 100$), so both models train under full masking for ${\approx}70\%$ of their steps.}

\noindent\textbf{Evaluation Benchmarks.}
\label{sec:benchmarks}
We evaluate on one ID benchmark, \textbf{COSMIC}~\cite{sikarwar2026communicating}, built on Infinigen scenes from our training domain, and five real-world OOD benchmarks covering diverse cross-view spatial reasoning skills.
 \textbf{MMSI-Bench}~\cite{yang2026mmsibench} poses expert-authored questions requiring spatial reasoning across multiple images of a scene (overall split). \textbf{MindCube}~\cite{wang2026mindcube} tests building a coherent spatial mental model from partial, incrementally revealed views (MindCube-Tiny, $1{,}050$ samples). \textbf{OmniSpatial}~\cite{jia2026omnispatial} covers higher-order relational reasoning and non-egocentric viewpoints (Complex Logic and Perspective Taking subsets, averaged). \textbf{STARE-Perspective}~\cite{li2025unfoldingspatialcognitionevaluating} requires reasoning about object relations from a viewpoint other than the camera's. \textbf{BLINK-MultiView}~\cite{fu2024blink} tests integrating evidence across multiple images of one scene. Per-benchmark details are in Appx.~\ref{sec:appendix_benchmarks}.

\vspace{-2mm}
\subsection{How to Imagine? Does View Dropout Make Visual Thinking Matter?}
\label{sec:does_pvam_help}
\vspace{-2mm}
We now test whether VDrop converts the generated thinking-image into a load-bearing component of reasoning, as designed in §\ref{sec:vdrop}.

\begin{table*}[t]
\centering
\adjustbox{max width=\textwidth}{%
\begin{tabular}{lccccc|ccccccc}
\toprule[1.5pt]
\textbf{Setting}
& \multicolumn{4}{c}{\textbf{COSMIC (in-domain)}}
& \cellcolor{avgcol}\textbf{Avg.}
& \textbf{MMSI}
& \textbf{MindCube}
& \multicolumn{2}{c}{\textbf{OmniSpatial}}
& \textbf{STARE}
& \textbf{BLINK}
& \cellcolor{avgcol}\textbf{Avg.} \\
\cmidrule(lr){2-5} \cmidrule(lr){9-10}
& \textbf{Anchor} & \textbf{Count.} & \textbf{Rel-Dist.} & \textbf{Rel-Dir.}
& \cellcolor{avgcol}\textbf{ID}
& \textbf{Overall} & \textbf{Overall}
& \textbf{CL} & \textbf{PT}
& \textbf{Persp.}
& \textbf{MultiView}
& \cellcolor{avgcol}\textbf{OOD} \\
\midrule
SenseNova-U1 (no FT)
& 68.8 & 31.2 & 44.8 & 30.8 & \cellcolor{avgcol}43.9
& 29.0 & \textbf{44.8} & 27.0 & 33.5 & \textbf{33.2} & 38.4 & \cellcolor{avgcol}35.1 \\
Panoramic SFT
& \textbf{72.8} & \textbf{48.4} & 48.8 & 50.0 & \cellcolor{avgcol}\textbf{55.0}
& 28.6 & 40.6 & 26.2 & 42.4 & 30.0 & 37.6 & \cellcolor{avgcol}34.2 \\
\;\;+ VDrop
& 66.8 & 45.6 & \textbf{49.2} & \textbf{52.4} & \cellcolor{avgcol}53.5
& \textbf{31.0} & 42.3 & \textbf{29.4} & \textbf{45.1} & 31.6 & \textbf{41.4} & \cellcolor{avgcol}\textbf{36.7}\,\textsuperscript{\textcolor{teal}{(+2.5)}} \\
\bottomrule[1.5pt]
\end{tabular}%
}
\caption{\textbf{Cross-architecture validation on SenseNova-U1.}
Panoramic visual thinking trained with standard SFT vs.\ with VDrop. The two blue \textbf{Avg.}\ columns report the in-domain mean over the four COSMIC subtasks and the out-of-domain mean over five real-world benchmarks. Best per column in bold.}
\label{tab:sensenova}
\vspace{-5mm}
\end{table*}

\noindent
\textbf{Impact of view dropout and visual thinking.}
\label{sec:vdrop_with_without}
Table~\ref{tab:vdrop_with_without} contrasts three visual-thinking strategies trained under standard SFT (§\ref{sec:umm}) against the same strategies trained with VDrop.\footnote{OmniSpatial reports two splits (CL and PT); we average them into a single score that contributes equally with the other four benchmarks to the out-of-domain mean.}
\textbf{(1)} \emph{VDrop consistently improves OOD cross-view spatial reasoning across all three thinking-image types:}
Adding VDrop raises the OOD average for every strategy: panoramic $37.6 \rightarrow 40.0$ ($+2.4${; significant under a paired bootstrap, $p=0.018$, Appx.~\ref{sec:appendix_significance}}), top-down $37.3 \rightarrow 38.0$ ($+0.7$), and point matching $35.2 \rightarrow 36.1$ ($+0.9$). The effect is positive across all three, with panoramic benefiting most. With VDrop, panoramic visual thinking reaches an OOD average of $40.0\%$, a $6.7$-point gain over vanilla BAGEL ($33.3\%$) and above Qwen3-VL-8B ($37.0\%$), a strong general-purpose VLM.
\textbf{(2)} \emph{With only 8K training samples, VDrop outperforms prior methods trained on far more data:}
Panoramic visual thinking with VDrop ($40.0\%$ OOD, 8K samples) outperforms prior BAGEL-based visual-thinking methods despite their much larger fine-tuning sets: ThinkMorph ($37.2\%$, $3\times$ the samples) and BAGEL-Zebra-CoT ($26.8\%$, $23\times$ the samples). Since these methods fine-tune the same BAGEL backbone, the results show that having the right intermediate representation and the training recipe (VDrop) matters more than the scale of fine-tuning data.
\textbf{(3)} \emph{Visual intermediates outperform non-visual baselines on both in- and out-of-domain:}
All three visual-thinking variants with VDrop beat both non-visual baselines on the ID and OOD averages. The OOD gains are substantial: relative to No-Think ($35.1\%$), panoramic gains $+4.9$ points {($p<0.001$, paired bootstrap; Appx.~\ref{sec:appendix_significance})}, top-down $+2.9$, and point matching $+1.0$, with larger gains over Text CoT ($+7.3$, $+5.3$, $+3.4$, respectively). On ID, the margin is smaller, as fine-tuning BAGEL on our 8K set already reaches $80.6\%$ without visual thinking (No-Think) and visual thinking lifts this only to $83$--$85\%$: the ID task is largely solved by fine-tuning alone, so we treat the OOD average, far from saturated, as the primary metric. 
Notably, Text CoT performs \emph{worse} than No-Think: a textual CoT hurts rather than helps. We attribute this to inconsistent textual supervision, as the same 3D scene admits many valid descriptions, making the generated traces often imprecise (Appx.~\ref{app:text_cot_anno}).
The VDrop hyperparameter ablation is in Appx.~\ref{sec:appendix_vdrop_ablation}.

\noindent\textbf{Does VDrop transfer to other UMM architectures?}
\label{sec:sensenova}
On SenseNova-U1, fine-tuned on the same spatial training data and evaluated on the same benchmarks, the headline finding mirrors BAGEL: VDrop again delivers the best OOD generalization (Table~\ref{tab:sensenova}).
However, Panoramic SFT alone brings no OOD gain here ($34.2\%$ vs.\ $35.1\%$ for the untrained model): SenseNova-U1 starts from a stronger base, and, as the underuse thesis predicts, SFT teaches the model to \emph{draw} the panorama but not to \emph{read} it.
VDrop closes this gap, raising the OOD average to $36.7\%$ ($+2.5$ over Panoramic SFT), with ID gains concentrated on the two relational subtypes that most require cross-view reasoning (Rel-Dir and Rel-Dist).
VDrop also outperforms a \emph{No-Think} control trained on the same data with the thinking-image removed ($35.5$ vs.\ $35.0$ on the OOD average excluding BLINK; Table~\ref{tab:sensenova_nothink}).\footnote{BLINK is excluded because the No-Think checkpoint collapses to a constant predictor there; see Appx.~\ref{sec:appendix_sensenova_nothink} for detailed analysis.}

\begin{table}[t]
\centering
\adjustbox{max width=\columnwidth}{%
\begin{tabular}{lcccccc}
\toprule[1.5pt]
\textbf{Setting}
& \textbf{MMSI}
& \textbf{MindCube}
& \multicolumn{2}{c}{\textbf{OmniSpatial}}
& \textbf{STARE}
& \cellcolor{avgcol}\textbf{Avg.} \\
\cmidrule(lr){4-5}
& \textbf{Overall} & \textbf{Overall}
& \textbf{CL} & \textbf{PT}
& \textbf{Persp.}
& \cellcolor{avgcol}\textbf{OOD} \\
\midrule
SenseNova-U1 (no FT)
& 29.0 & \textbf{44.8} & 27.0 & 33.5 & \textbf{33.2} & \cellcolor{avgcol}34.3 \\
No-Think
& \textbf{32.0} & 42.2 & 26.2 & \textbf{50.5} & 27.6 & \cellcolor{avgcol}35.0 \\
Panoramic SFT
& 28.6 & 40.6 & 26.2 & 42.4 & 30.0 & \cellcolor{avgcol}33.4 \\
\;\;+ VDrop
& 31.0 & 42.3 & \textbf{29.4} & 45.1 & 31.6 & \cellcolor{avgcol}\textbf{35.5} \\
\bottomrule[1.5pt]
\end{tabular}%
}
\caption{\textbf{No-Think control on SenseNova-U1 (OOD, BLINK excluded).}
Best per column in bold.}
\label{tab:sensenova_nothink}
\vspace{-5mm}
\end{table}

\noindent\textbf{Generate-then-Blind: Is the impact of thinking-image causal?}
\label{sec:generate_then_blind}
Performance gains alone do not imply that the answer \emph{causally} depends on the generated thinking-image: the model could produce a decorative image whose only effect is regularisation.
To test whether the thinking-image causally affects answer prediction, we introduce a \emph{generate-then-blind} intervention at inference. For each test item, the model first generates the thinking-image as in normal inference; before answer decoding begins, we then mask the answer span's attention to the thinking-image tokens. The answer span still attends to $V_1$, $V_2$, and the question, but can no longer read the thinking-image: the model still ``thinks'' by producing the image, yet answers without consulting it. A model that genuinely uses the thinking-image should lose accuracy under blinding; one that ignores it should be unaffected.

\begin{figure}[t]
    \centering
    \includegraphics[width=0.85\linewidth]{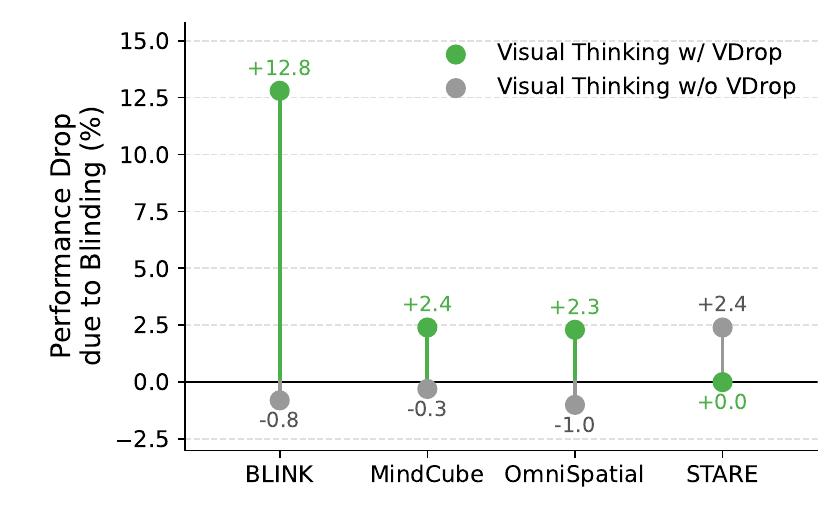}
\caption{\textbf{Generate-then-blind probe across 4 OOD benchmarks.} 
Accuracy drop when the generated thinking-image is blinded at answer time; a larger drop means more dependence on the thinking-image. VDrop-trained models show larger drops on three benchmarks.
}
    \label{fig:generate_then_blind}
    \vspace{-5mm}
\end{figure}

We apply this probe to two panoramic-trained variants, standard visual-thinking SFT and SFT with VDrop (Figure~\ref{fig:generate_then_blind}; full setup in Appx.~\ref{sec:appendix_generate_then_blind}). 
The VDrop variant shows substantial accuracy drops under blinding on most OOD benchmarks while standard SFT is largely invariant, confirming that VDrop makes the thinking-image load-bearing rather than decorative.
A per-category breakdown on MMSI (Figure~\ref{fig:generate_then_blind_mmsi} in Appx.~\ref{sec:appendix_generate_then_blind}) shows VDrop's blinding effect concentrates on questions answerable by visually aligning the two views, confirming its causal dependence on the thinking-image appears where visual reasoning helps.

\begin{figure}[t]
    \centering
    \includegraphics[width=0.95\linewidth]{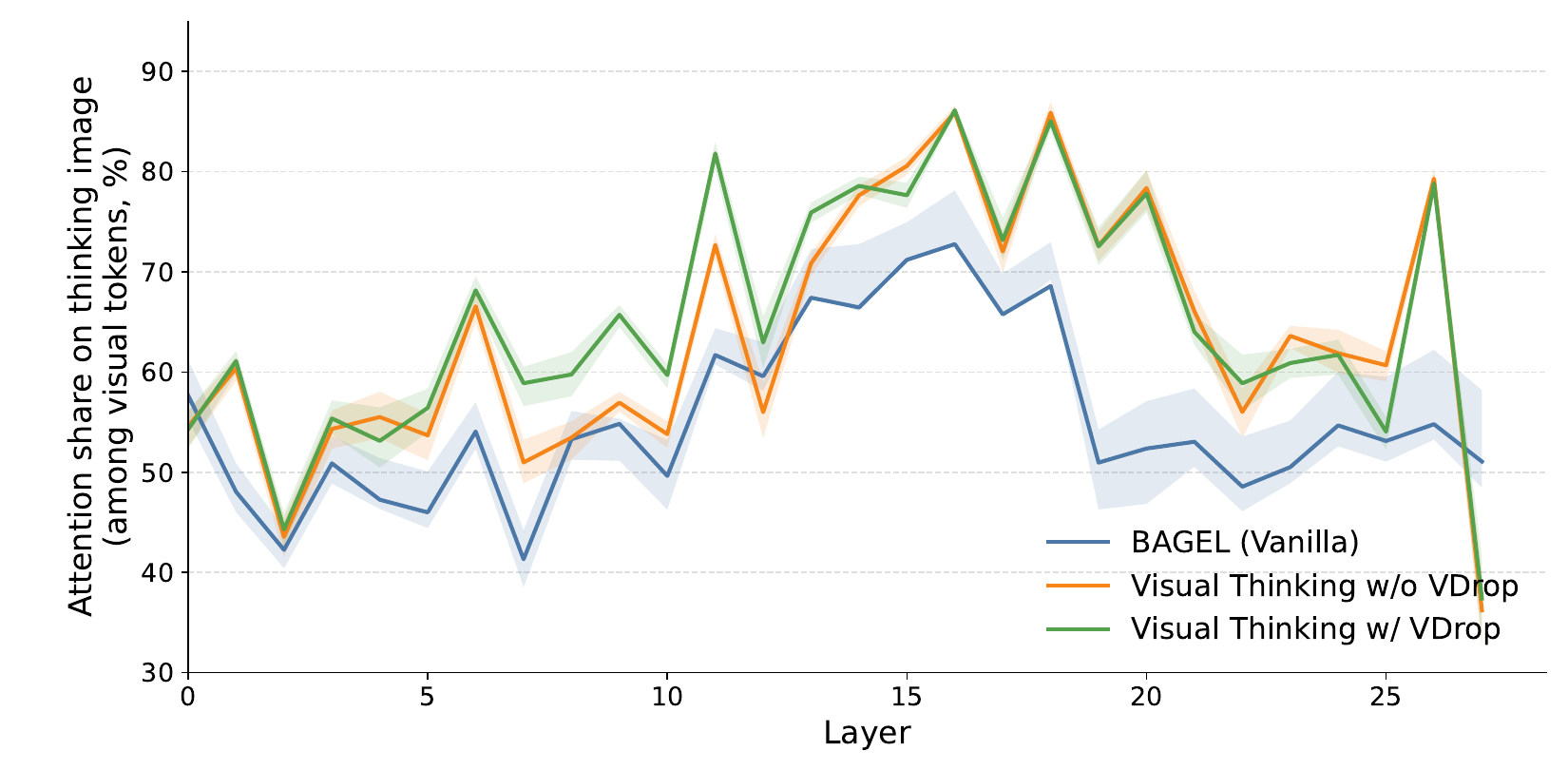}
    \caption{\textbf{Mean answer-token attention on thinking-image tokens across decoder layers (BLINK).} VDrop-trained model places more attention on the generated thinking-image than standard SFT model, especially in early and mid layers. Shaded bands show the interquartile range across samples (middle 50\%).}
    \label{fig:vt_share_layers}
    \vspace{-5mm}
\end{figure}

\noindent
\textbf{Mechanism check: Does VDrop increase answer-token attention to the thinking-image?}
\label{sec:vdrop_attention}
We measure how the answer tokens divide their attention among the visual inputs during answer generation. At each decoder layer, we compute how much the answer attends to the generated thinking-image as a fraction of its total attention to all visual content, i.e.\ the thinking-image plus the two input views $V_1, V_2$. A higher fraction means the answer relies more on the thinking-image than on the input views (full setup in Appx.~\ref{sec:appendix_attention_probe}).\footnote{Vanilla BAGEL does not reliably emit a thinking-image; we force one by injecting the image-generation token, giving all three models a comparable thinking-image span.}
Averaging across all BLINK samples and decoder layers (Figure~\ref{fig:vt_share_layers}), the thinking-image's share of the answer span's visual attention rises across the three models: $55.3\%$ for vanilla BAGEL, $63.6\%$ for standard SFT, and $65.2\%$ with VDrop. The gap widens in early and mid decoder layers, where VDrop attends $3.9$ percentage points more to the thinking-image than standard SFT.\footnote{The three shares average over all decoder layers; the $3.9$-point figure averages over the first $14$ layers, where the effect concentrates.} We observe the same early-to-mid-layer pattern on STARE (Appx.~\ref{sec:appendix_attention_probe}).

\finding{1}{
Visual thinking improves cross-view spatial reasoning: generating an intermediate thinking-image outperforms non-visual baselines on both ID and OOD benchmarks.
}

\finding{2}{
VDrop consistently improves cross-view spatial reasoning across all three thinking-image types and across two architecturally distinct UMMs (BAGEL and SenseNova U1), increasing the answer span's attention to the thinking-image and making it a load-bearing component of reasoning.
}

\subsection{What to Imagine? A Learnability -- Informativeness Tradeoff}
\label{sec:l_i_tradeoff}
With VDrop ensuring the thinking-image is genuinely used, we now ask which thinking-image type yields the largest gains. We formalise the end-to-end benefit of a type $T$ along two axes.

\noindent\textbullet~\textbf{Informativeness} $I(T)$: how much a \emph{perfect} instance of $T$ would reduce the model's reasoning burden on the target task. A panorama exposes the joint scene directly and a top-down view exposes the global layout, whereas point matching only marks correspondences within the original views and adds no new viewpoint.

\noindent\textbullet~\textbf{Learnability} $L(T)$: how reliably a UMM can be trained to produce faithful instances of $T$ from the available data. A panorama is a single coherent rendering the model can generate directly, whereas a faithful top-down view requires synthesising an unseen viewpoint, and precise point markers must be placed exactly on small target objects.

We hypothesise that the benefit of visual thinking is bounded by these two axes: a thinking-image type helps only if it is both informative \emph{and} learnable, with neither alone sufficient.
{This choice of axes is grounded in imitation learning: visual-thinking SFT trains the model to reproduce a target thinking-image, i.e., behavioural cloning of the target~\cite{stiennon2020learning,ouyang2022training}, whose theory bounds the cloned policy by two independent error terms: \textit{(i)} the value of the imitated target, since cloning is at best as good as the target it imitates~\cite{rajaraman2020toward}, and \textit{(ii)} the realizability of that target under the learner, since a target the model class cannot represent leaves an irreducible misspecification error~\cite{espinosa2025efficient}.
These correspond exactly to $I(T)$ and $L(T)$ (details in Appx.~\ref{sec:appendix_li_grounding}).}
{We measure each axis in isolation on a paired subset of the COSMIC test set\footnote{The $L$--$I$ analysis uses the paired subset of $n{=}720$ COSMIC test examples for which ground-truth thinking-images of all strategies could be rendered (requiring complete source-mesh data); the subset spans the four subtypes near-uniformly.} by controlling \emph{which} thinking-image the answerer sees: ground-truth renders isolate the target's value from generation quality, while generated images expose how much of that value the model realises; any gap between the two measurements is the model failing to realise the target.}

\noindent\textbf{Informativeness: oracle measurement.} 
We feed \emph{ground-truth} renders of each thinking-image type as a third image to two off-the-shelf VLMs, Qwen3-VL-32B and Qwen3-VL-235B (Table~\ref{tab:two_reader_oracle}). 
The informativeness ranking, panorama $>$ top-down $>$ point matching, is consistent across both VLMs. 
\emph{Panorama} delivers the largest uplift, concentrated on the relative-distance and relative-direction subtypes whose answers require a viewpoint neither input view provides. \emph{Top-down} yields a smaller but consistently positive uplift, exposing global spatial relations through an allocentric view. \emph{Point matching} is flat-to-negative on both: its annotations stay in the original viewpoint and add little geometric structure beyond the input views.
{Anchor moves little under \emph{any} intermediate: it is already near-saturated from the input views alone (the Qwen3-VL-32B answerer reaches $83.33$ on Anchor with just $V_1$ and $V_2$), leaving little headroom for a third image, whereas the other subtypes start far lower and gain much more.
This dip is a property of the probe, not of the trained UMM: end-to-end, the panoramic variants still beat direct answering on Anchor ($93.6$/$89.2$ vs.\ $86.8$; Table~\ref{tab:vdrop_with_without}).}

\begin{table}[t]
\centering
\small
\adjustbox{max width=\linewidth}{%
\begin{tabular}{lrrrrr}
\toprule
 \textbf{Condition} & \textbf{Anchor} & \textbf{Counting} & \textbf{Rel-Dist} & \textbf{Rel-Dir} & \textbf{Overall} \\
 \midrule
\multicolumn{6}{l}{\textit{Qwen3-VL-32B-Instruct}} \\
\midrule
Input views only  & 83.33 & 73.22 & 56.32 & 39.66 & 62.78 \\
  $T_{\mathrm{pano}}$ ($\Delta$)        & $-5.95$ & \best{${+1.64}$} & \best{$\mathbf{+12.63}$} & \best{$\mathbf{+25.14}$} & \best{$\mathbf{+8.61}$} \\
     $T_{\mathrm{topdown}}$ ($\Delta$) & $-5.36$ & \best{$\mathbf{+3.83}$} & \best{$+5.26$} & \best{$+13.97$} & \best{$+4.58$} \\
   $T_{\mathrm{point\,mat.}}$ ($\Delta$)         & \best{$\mathbf{+2.98}$} & $-10.38$ & $-0.53$ & \best{$+2.23$} & $-1.53$ \\
\midrule
\multicolumn{6}{l}{\textit{Qwen3-VL-235B-A22B-Instruct}} \\
\midrule
Input views only  & 73.21 & 66.12 & 49.47 & 39.11 & 56.67 \\
  $T_{\mathrm{pano}}$ ($\Delta$)        & \best{$\mathbf{+0.60}$}         & \best{$\mathbf{+9.84}$} & \best{$\mathbf{+15.26}$ }& \best{$\mathbf{+16.76}$} & \best{$\mathbf{+10.83}$}\\
   $T_{\mathrm{topdown}}$ ($\Delta$) & $-6.54$ & $-1.09$ & \best{$+10.00$} & \best{$+8.38$} & \best{$+2.92$}  \\
   $T_{\mathrm{point\,mat.}}$ ($\Delta$)         & $\phantom{+}0.00$ & $-9.29$         & $\phantom{+}0.00$ & \best{$+6.15$}         & $-0.83$ \\
\bottomrule
\end{tabular}%
}
\caption{\textbf{Oracle measurement of $I(T)$ on COSMIC}. For each VLM block, the \emph{Input views only} row reports accuracy (\%) given just $V_1$ and $V_2$; subsequent rows report uplift $\Delta$ over that baseline when a third image of type $T$ is supplied. }
\label{tab:two_reader_oracle}
\vspace{-5mm}
\end{table}

\noindent\textbf{Learnability: generation quality.} 
We assess whether BAGEL can produce faithful instances of each type, using two measurements computed on the \emph{same} set of BAGEL-generated thinking-images, so generation fidelity and functional usefulness are evaluated on identical images. 
\textbf{\emph{Directly}}, we compute SigLIP cosine similarity between BAGEL-generated thinking-images and their ground-truth renders. Panorama and top-down generations are the most faithful, with near-identical similarity ($0.950$ and $0.948$), both above point matching ($0.928$). 
But SigLIP captures whole-image appearance, not geometric fidelity: a generated top-down view can resemble its target room while misplacing objects or distorting depth (see Appendix~\ref{sec:appendix_qualitative} for qualitative examples).
We therefore treat it as a necessary but insufficient signal, and turn to a functional measurement. \textbf{\emph{Indirectly}}, we feed BAGEL's \emph{generated} thinking-image to a frozen VLM (Qwen3-VL-235B) and measure whether it still helps answer cross-view questions (Table~\ref{tab:generation_understanding}): a learnable strategy should yield images the VLM can use as spatial evidence. Generated panoramas remain net-positive on all three spatial subtypes (Counting, Rel-Dist, Rel-Dir), showing BAGEL transfers their spatial \emph{structure} even from generated images; the lone regression is on Anchor, which benefits little even at the oracle level. Generated top-down images help on two spatial subtypes but regress on the others, leaving overall accuracy below baseline, so BAGEL captures the top-down layout only partially. Point matching is flat-to-negative throughout, consistent with its low oracle informativeness rather than a generation failure.

\begin{table}[t]
\centering
\small
\setlength{\tabcolsep}{4pt}
\adjustbox{max width=\linewidth}{%
\begin{tabular}{lrrrrr}
\toprule[1.5pt]
\textbf{Condition} & \textbf{Anchor} & \textbf{Counting} & \textbf{Rel-Dist} & \textbf{Rel-Dir} & \textbf{Overall} \\
\midrule
Input views only  & 73.21 & 66.12 & 49.47 & 39.11 & 56.67 \\
$T_{\mathrm{pano\_gen}}$ ($\Delta$)
& $-13.69$
& \best{$\mathbf{+3.28}$}
& \best{$+2.63$}
& \best{$\mathbf{+6.70}$}
& \best{$\phantom{+}0.00$} \\
$T_{\mathrm{td\_gen}}$ ($\Delta$)
  & $-13.10$ & $-2.19$ &\textbf{ \best{$\mathbf{+8.42}$}} & \best{$+2.79$} & $-0.69$ 
\\
$T_{\mathrm{point\,mat.\_gen}}$ ($\Delta$)
& $-0.60$
& $-2.19$
& $-3.68$
& $-0.56$
& $-1.81$ \\
\bottomrule[1.5pt]
\end{tabular}%
}
\caption{\textbf{Generated-image measurement of $L(T)$ on COSMIC.}
The \emph{Input views only} row reports accuracy (\%) given $V_1$ and $V_2$ on Qwen3-VL-235B-A22B-Instruct; subsequent rows report the uplift $\Delta$ from supplying a thinking-image of type $T$, generated by BAGEL after VDrop training, as a third image.
}
\label{tab:generation_understanding}
\vspace{-5mm}
\end{table}

\finding{3}{On the L–I framework, top-down is informative but only partially learnable, while point matching is limited mainly by low informativeness and only moderately learnable, as its fine markers are hard to place precisely. Panorama alone is strong on both axes.}

\section{Conclusion}
\label{sec:conclusion}
We studied how to make visual thinking matter for cross-view spatial reasoning in UMMs, and which kind of visual thinking is most effective. We identified \emph{under-use} as a pervasive failure of current visual-thinking pipelines and proposed \textbf{View Dropout (VDrop)}, a training-time intervention that forces the generated thinking-image to become a load-bearing component of reasoning. With visual thinking made load-bearing, we framed the choice of intermediate as a \emph{Learnability–Informativeness tradeoff} and identified panoramic visual thinking with VDrop as the only configuration that is simultaneously informative, learnable, and used. Trained on only 8K samples, it achieves the best out-of-distribution generalization, outperforming prior methods on the same backbone trained on at least $3\times$ more data. The bottleneck for visual thinking is not data scale, but a training signal that forces the thinking-image to be used.      
\clearpage
\newpage

\section*{Limitations}


First, our study focuses on a single task family, cross-view spatial reasoning, where text-based reasoning demonstrably fails and visual imagination is inherently required; whether VDrop and the Learnability--Informativeness framework extend to other visual-thinking tasks remains to be tested.
Second, VDrop makes the thinking-image causally load-bearing, but it does not by itself improve the \emph{quality} of the generated thinking-image: when the generated image is low-fidelity or not genuinely useful for the question, forcing the answer to route through it provides little benefit. 
VDrop is therefore complementary to, not a substitute for, supervision that improves thinking-image generation itself; combining VDrop with higher-fidelity thinking-image targets is a natural direction for future work.
\section*{Acknowledgements}
We thank the Mila IDT team for providing the computational resources for this research, their technical support and for maintaining the Mila compute cluster. 
We also thank Calcul Québec and Digital Research Alliance of Canada for providing the computational resources for this research. 
Throughout this project, Aishwarya Agrawal received support from the Canada CIFAR AI Chair
award.



\bibliography{custom,references}
\clearpage
\newpage
\appendix

\appendix
\section*{Appendix}

\phantomsection
\label{app:overview}

\vspace{-0.5em}
\begin{tcolorbox}[
    enhanced,
    colback=white,
    colframe=teal!60!black,
    coltitle=white,
    colbacktitle=teal!60!black,
    title=\textbf{Appendix Overview},
    fonttitle=\bfseries,
    attach boxed title to top left={xshift=6pt, yshift=-\tcboxedtitleheight/2},
    boxed title style={
        colframe=teal!60!black,
        colback=teal!60!black,
        arc=2pt,
        boxrule=0.4pt,
        left=4pt, right=4pt, top=1pt, bottom=1pt,
    },
    arc=2pt,
    boxsep=2pt,
    left=6pt, right=6pt, top=8pt, bottom=3pt,
    boxrule=0.8pt,
    drop fuzzy shadow=black!25,
    before skip=10pt, after skip=4pt,
    breakable
]
\small
\renewcommand{\arraystretch}{1.4}
\begin{tabular}{@{}p{0.06\textwidth}p{0.88\textwidth}@{}}

\hyperref[sec:appendix_benchmarks]{\textbf{\textcolor{teal!60!black}{\ref*{sec:appendix_benchmarks}}}} & \textbf{Evaluation Benchmarks}\\

\hyperref[sec:appendix_training_data]{\textbf{\textcolor{teal!60!black}{\ref*{sec:appendix_training_data}}}} & \textbf{Training Data}\\

\hyperref[sec:appendix_experiments]{\textbf{\textcolor{teal!60!black}{\ref*{sec:appendix_experiments}}}} & \textbf{Experiments}\\

\end{tabular}
\end{tcolorbox}
\vspace{1em}

\section{Evaluation Benchmarks}
\label{sec:appendix_benchmarks}

We evaluate on one in-domain (ID) benchmark and five real-world out-of-domain (OOD) benchmarks covering diverse cross-view spatial reasoning skills. All benchmarks use multiple-choice questions, and we report accuracy. Unless otherwise specified, we parse each model output to extract the predicted answer token and apply exact-match scoring against the ground-truth answer.
\footnote{The artifacts used in this work, including the base model and associated resources, are released under the Apache-2.0 license. Our use is limited to research on cross-view spatial reasoning and follows the artifacts' stated terms and intended use.}

\paragraph{ID Benchmark.}
\textbf{COSMIC}~\cite{sikarwar2026communicating} is a cross-view spatial reasoning benchmark built on Infinigen-generated scenes, making it in-domain with respect to our training data.
COSMIC covers two levels of spatial reasoning. Object-level tasks include \emph{anchor recognition} (identifying objects shared across views) and \emph{global counting} (aggregating object instances across views while correctly disambiguating shared and unique instances).
Relation-level tasks include \emph{relative distance} (inferring which object is closest or farthest from a target distributed across views) and \emph{relative direction} (inferring the egocentric direction of a target object absent from the answerer's view, requiring cross-view perspective transformation).
Each subtask contains 250 samples, with no scene overlap between the evaluation set and our training samples.

\paragraph{OOD Benchmarks.}
We evaluate OOD generalisation on five benchmarks covering real-world environments and diverse spatial reasoning skills.

\textbf{MMSI-Bench}~\cite{yang2026mmsibench} is a multi-image spatial intelligence benchmark containing 1{,}000 challenging multiple-choice questions, each meticulously crafted by six 3D-vision researchers and paired with carefully designed distractors and a stepwise reasoning process.
MMSI-Bench is highly challenging: the strongest open-source models achieve roughly 30\% accuracy and GPT-5 reaches 41.9\%, while humans score 97.2\%.
We report the \emph{Overall} accuracy across all MMSI question types as the headline MMSI score.
Following the official recommendation, we use Gemini-3.0-Flash to extract the predicted answer from each model output, and then apply exact-match scoring between the extracted answer and the ground-truth answer for each multiple-choice question.

\textbf{MindCube}~\cite{wang2026mindcube} tests whether models can build spatial mental models from partial observations.
It evaluates three core spatial settings: \emph{Rotation} (interpreting multiple orthogonal views from a static observation point, requiring holistic understanding despite incremental visibility shifts), \emph{Around} (leveraging occlusion to test object permanence and the ability to convert lateral relations in frontal views into depth cues in side views), and \emph{Among} (maintaining spatial consistency across views captured around a central object, requiring models to deduce overall spatial arrangement when not all elements are simultaneously visible).
We evaluate on MindCube-Tiny, a stratified subset of 1{,}050 samples balanced across the three settings.

\textbf{OmniSpatial}~\cite{jia2026omnispatial} is a comprehensive spatial reasoning benchmark for VLMs covering a broad set of spatial skills.
We evaluate on two subsets relevant to cross-view spatial reasoning: \emph{Complex Logic} (CL), which involves higher-order reasoning about relations, transformations, and geometric structure; and \emph{Perspective Taking} (PT), which probes the ability to reason about a scene from a non-egocentric viewpoint.
We report each subset individually in the main results table and an unweighted average of the two as the OmniSpatial headline score.

\textbf{STARE-Perspective}~\cite{li2025unfoldingspatialcognitionevaluating} is the perspective-taking split of the STARE benchmark, which evaluates a model's ability to reason about object positions and relations relative to a non-egocentric viewpoint specified in the question.

\textbf{BLINK-MultiView}~\cite{fu2024blink} is the multi-view split of BLINK, which tests visual reasoning that requires integrating evidence across multiple images of the same scene rather than from a single image alone.

\section{Training Data Details}
\label{sec:appendix_training_data}

The synthetic Infinigen source is chosen for two reasons: (i) procedural rendering provides ground-truth thinking-images for every variant in §\ref{sec:strategies} (top-down maps, panoramic stitches, and point-matching overlays) free of real-world rendering noise, and (ii) full access to object-level 3D annotations (positions, bounding boxes, categories) lets us automatically construct cross-view spatial questions whose answers are unambiguous.

\paragraph{Scene generation.}
Infinigen Indoors \cite{infinigen} generates diverse photorealistic indoor scenes, each populated with a structured layout of furniture, architectural elements, and everyday objects.
For every scene we render two egocentric views with deliberately overlapping fields of view: the two camera poses are sampled such that a portion of the scene is co-visible across views, ensuring that shared objects or shared regions anchor the two viewpoints and make a layout connection between them feasible.
Alongside each view pair, we render the top-down bird's-eye-views, stitched panoramic views, and point-matching overlays used as ground-truth $I_{\text{vt}}$ for the variants in §\ref{sec:strategies}.

\paragraph{Question types.}
We automatically generate four types of cross-view spatial questions, each targeting a distinct spatial skill:
\begin{itemize}
    \item \textbf{Anchor.} Given two cross-view images, identify the common object(s). This tests cross-view correspondence: the model must match object identity despite viewpoint-induced appearance changes such as occlusion, scale, and aspect-ratio shifts.

    \item \textbf{Counting.} Given two cross-view images, count the total number of instances of a specified object category. This requires both cross-view correspondence and enumeration: the model must count instances in each view and resolve which are shared to avoid double-counting.

    \item \textbf{Relative Distance.} Given two cross-view images and a set of target objects, determine which object is farthest from a reference object. This requires 3D metric layout recovery, estimating inter-object distances from 2D projections.

    \item \textbf{Relative Direction.} Given two cross-view images, determine the direction of an object visible in one image relative to the viewpoint of the other. This requires reference-frame transformation: localising an object in 3D from one view and projecting its direction into the other view's coordinate frame.
\end{itemize}

\paragraph{Training data distribution.}
Table~\ref{tab:training_data} reports the per-type sample and unique-scene counts for the $8$K Infinigen training set. We deliberately allocate fewer samples to Anchor and Counting than to the relative-distance and relative-direction types. Anchor recognition is implicitly exercised whenever the model answers any cross-view question, and Counting concerns objects largely visible within the overlapping region of the two views rather than genuine cross-view spatial relations. Both are reflected in the higher zero-shot accuracy of off-the-shelf VLMs on Anchor and Counting than on the relational subtypes (Table~\ref{tab:two_reader_oracle}). We therefore reallocate sample budget toward relative distance and relative direction, the harder subtypes, while retaining sufficient accuracy on Anchor and Counting at the reduced counts.

\begin{table}[t]
\centering
\small
\adjustbox{max width=\linewidth}{%
\begin{tabular}{lrr}
\toprule[1.5pt]
\textbf{Question Type} & \textbf{Samples} & \textbf{Unique Scenes} \\
\midrule
Anchor                  &   730 &   455 \\
Counting                & 1{,}191 &   854 \\
Relative Distance       & 3{,}000 &   192 \\
Relative Direction      & 3{,}000 &   893 \\
\midrule
\textit{Total}          & \textit{7{,}921} & \textit{1{,}584} \\
\bottomrule[1.5pt]
\end{tabular}
}
\caption{Training data distribution by question type. All training samples are procedurally rendered from Infinigen Indoors~\cite{infinigen}; no external real-world data is used.}
\label{tab:training_data}
\end{table}

\subsection{Text Chain-of-Thought Annotation}
\label{app:text_cot_anno}
Each training sample comes with a ground-truth multiple-choice answer derived from the underlying 3D scene, but no natural-language rationale. To produce one, we prompt an off-the-shelf large multimodal model, Qwen3-VL-235B-A22B-Instruct, conditioning it on the two camera images and a category-specific textual prompt. The prompt has three parts: a short instruction block specifying the reasoning shape for that question category, one curated in-context example, and a scene-metadata block synthesised from the QA row (the per-scene list of visible objects with short descriptions, and the two camera poses).

A central design choice is an \emph{oracle--trace separation}. The raw numeric quantities used to construct the question (angles, distances, exact object counts, and the gold answer letter) are passed to the annotator as a tagged private ``cheat sheet'', so that it lands on the correct option with high probability; the system prompt, however, forbids citing these quantities in the rationale. The trace must instead argue from visual cues a student model could verify from the images alone: qualitative placements (e.g.\ mid-centre-right), foreground/background contrasts, direction words, and, for view-dependent categories, shared landmarks together with a relative-camera-pose hint that lets the trace argue which side an off-screen target lies on.

We use an open-source model rather than a proprietary one such as Gemini for two reasons. Each of the 8K samples requires conditioning on two input images and a long oracle-information context, so annotating the full set with a proprietary API would be costly; an open-source annotator also keeps the pipeline fully reproducible. Despite the oracle-conditioning protocol, the synthesised text traces are not consistently high-quality: producing a linguistically consistent description of a 3D scene is difficult, since the same spatial configuration admits many equally valid descriptions, yielding surface-form inconsistency across training samples. Visual thinking sidesteps this: a ground-truth panoramic or top-down rendering is a deterministic function of the scene, so the supervision is consistent by construction. This may partially explain why Text CoT underperforms even No-Think in our end-to-end results (§\ref{sec:vdrop_with_without}). Designing higher-quality interleaved reasoning supervision that combines visual and textual chains-of-thought is a promising direction for future work.

\section{Experiments}
\label{sec:appendix_experiments}

\subsection{Confidence Intervals and Significance Tests}
\label{sec:appendix_significance}

{We assess the robustness of the cross-strategy ranking in Table~\ref{tab:vdrop_with_without} with bootstrap resampling over test items.}

\paragraph{Paired significance tests.}
For each pair of conditions, we run a \emph{paired} bootstrap over per-item accuracy differences: within each benchmark we resample its questions with replacement and recompute the accuracy difference between the two conditions on the same resampled items, then average the five benchmark-level differences with equal weight, following the OOD protocol of Table~\ref{tab:vdrop_with_without} (OmniSpatial's CL and PT splits are first combined into a single OmniSpatial score).
Both key comparisons are significant: Panoramic~+~VDrop beats No-Think on the OOD average by $+4.9$ points ($95\%$ CI $[+2.5, +7.1]$, $p < 0.001$), and beats Panoramic without VDrop by $+2.4$ points ($95\%$ CI $[+0.5, +4.7]$, $p = 0.018$).
The pairing removes item-level variance shared between conditions, which is appropriate here because all conditions are evaluated on identical test items.

\paragraph{Per-condition confidence intervals.}
Table~\ref{tab:bootstrap_ci} reports per-condition $95\%$ percentile-bootstrap confidence intervals ($10{,}000$ resamples, resampling items) for every in-domain subtype and OOD benchmark.
The VDrop interval on the OOD average, $[38.1, 42.4]$, does not overlap the No-Think interval $[33.0, 37.2]$ and only marginally overlaps the Panoramic-without-VDrop interval $[35.5, 39.9]$; the paired tests above, which account for the shared items, confirm both gaps are significant.

\begin{table*}[t]
\centering
\adjustbox{max width=\textwidth}{%
\begin{tabular}{lcccc|ccccccc}
\toprule[1.5pt]
\textbf{Condition}
& \multicolumn{4}{c}{\textbf{COSMIC (in-domain)}}
& \textbf{MMSI}
& \textbf{MindCube}
& \multicolumn{2}{c}{\textbf{OmniSpatial}}
& \textbf{STARE}
& \textbf{BLINK}
& \cellcolor{avgcol}\textbf{Avg.} \\
\cmidrule(lr){2-5} \cmidrule(lr){6-11}
& \textbf{Anchor} & \textbf{Count.} & \textbf{Rel-Dist.} & \textbf{Rel-Dir.}
& \textbf{Overall} & \textbf{Overall}
& \textbf{CL} & \textbf{PT}
& \textbf{Persp.}
& \textbf{MultiView}
& \cellcolor{avgcol}\textbf{OOD} \\
\midrule
No-Think
& 86.8 {\scriptsize[82.4, 90.8]} & 82.4 {\scriptsize[77.6, 86.8]} & 67.6 {\scriptsize[61.6, 73.6]} & 85.6 {\scriptsize[81.2, 89.6]}
& 27.4 {\scriptsize[24.7, 30.2]} & 41.1 {\scriptsize[38.1, 44.0]} & 30.2 {\scriptsize[24.6, 36.1]} & 44.0 {\scriptsize[39.9, 48.1]}
& 24.8 {\scriptsize[19.6, 30.4]} & 45.1 {\scriptsize[36.8, 53.4]} & \cellcolor{avgcol}35.1 {\scriptsize[33.0, 37.2]} \\
Text CoT
& 56.0 {\scriptsize[49.6, 62.0]} & 60.4 {\scriptsize[54.4, 66.4]} & 44.4 {\scriptsize[38.4, 50.4]} & 40.8 {\scriptsize[34.8, 47.2]}
& 24.6 {\scriptsize[22.0, 27.3]} & 25.3 {\scriptsize[22.8, 28.0]} & 29.4 {\scriptsize[23.8, 34.9]} & 35.8 {\scriptsize[31.9, 39.9]}
& 26.0 {\scriptsize[20.8, 31.6]} & 54.9 {\scriptsize[46.6, 63.2]} & \cellcolor{avgcol}32.7 {\scriptsize[30.4, 34.7]} \\
Panoramic View
& 93.6 {\scriptsize[90.4, 96.4]} & 83.6 {\scriptsize[78.8, 88.0]} & 76.4 {\scriptsize[71.2, 81.6]} & 87.2 {\scriptsize[83.2, 91.2]}
& 24.9 {\scriptsize[22.3, 27.6]} & 36.9 {\scriptsize[33.9, 39.8]} & 32.9 {\scriptsize[27.4, 38.9]} & 43.9 {\scriptsize[39.8, 48.0]}
& 32.4 {\scriptsize[26.8, 38.4]} & 55.6 {\scriptsize[47.4, 63.9]} & \cellcolor{avgcol}37.6 {\scriptsize[35.5, 39.9]} \\
\;\;+ VDrop
& 89.2 {\scriptsize[85.2, 92.8]} & 78.8 {\scriptsize[73.6, 83.6]} & 74.8 {\scriptsize[69.2, 80.0]} & 93.2 {\scriptsize[90.0, 96.0]}
& 26.0 {\scriptsize[23.3, 28.7]} & 34.1 {\scriptsize[31.2, 37.0]} & 38.9 {\scriptsize[32.9, 45.2]} & 45.3 {\scriptsize[41.2, 49.4]}
& 35.6 {\scriptsize[29.6, 41.6]} & 62.4 {\scriptsize[54.1, 70.7]} & \cellcolor{avgcol}\textbf{40.0} {\scriptsize[38.1, 42.4]} \\
\bottomrule[1.5pt]
\end{tabular}%
}
\caption{\textbf{Per-condition $95\%$ bootstrap confidence intervals} ($10{,}000$ resamples, resampling items within each benchmark or subtype).
The OOD \textbf{Avg.} is the equal-weight mean over the five OOD benchmarks; OmniSpatial's CL and PT splits are first combined into a single OmniSpatial score, so the average is over benchmarks, not over the displayed columns.
All four conditions fine-tune the same BAGEL backbone on the identical 8K training set and differ only in the thinking strategy.}
\label{tab:bootstrap_ci}
\vspace{-4mm}
\end{table*}

\subsection{VDrop Mask Design Ablation}
\label{sec:appendix_vdrop_ablation}

Having established in §\ref{sec:vdrop_with_without} that VDrop helps across thinking-image types, we ablate which axis of the mask is responsible for the gain. Holding the thinking-image type (Panoramic) and training recipe (LoRA SFT, 8K samples) fixed, we vary three axes of VDrop: the patch-selection \emph{Strategy}, \textbf{Region} (a contiguous bounding-box of the chosen view's patches) versus \textbf{Random} (an i.i.d.\ random subset at the same drop ratio); the masked \emph{Scope}, \textbf{One view} (a single primary view sampled uniformly from $\{V_1, V_2\}$) versus \textbf{Two views} (both input views masked together); and the \emph{drop ratio}. A no-mask reference (Panoramic without VDrop) and a no-warmup ablation (masking from step~0, no anneal) are reported for comparison. Table~\ref{tab:vdrop_ablation} reports aggregate ID and OOD accuracy for each variant.
The default configuration used throughout the paper, \textbf{Region} masking on a single view at $50\%$ drop with the warmup--anneal curriculum, achieves the best OOD accuracy ($40.0$). Four findings stand out. \textbf{(i) Region beats Random:} replacing the contiguous region with random patches at the same drop ratio costs $4.4$ OOD points, indicating that spatially coherent occlusion is what forces the thinking-image to encode localised structure. \textbf{(ii) One view beats two:} masking both views simultaneously costs $3.2$ OOD points, suggesting that retaining one full view as an anchor is necessary for the model to learn what content needs routing through $I_{\mathrm{vt}}$. \textbf{(iii) An intermediate drop ratio is best:} $50\%$ outperforms both $30\%$ ($+7.4$ OOD) and $80\%$ ($+3.5$ OOD); too little masking leaves the input-view shortcut intact, while too much removes spatial evidence the model needs. \textbf{(iv) Warmup is essential:} applying full masking from step~0 without the warmup--anneal schedule drops OOD accuracy by $6.3$ points at the same drop ratio, consistent with the curriculum motivation in §\ref{sec:vdrop}: the model must first learn what to put in $I_{\mathrm{vt}}$ before being forced to depend on it.

\begin{table}[t]
\centering
\small
\setlength{\tabcolsep}{4pt}
\adjustbox{max width=\columnwidth}{%
\begin{tabular}{llcrr}
\toprule[1.5pt]
\textbf{Strategy} & \textbf{Scope} & \textbf{Drop $\rho$ \%}
& \textbf{ID} & \textbf{OOD} \\
\midrule
\multicolumn{5}{l}{\textit{No VDrop (reference)}} \\
None   & n/a       & 0  & {85.2} & 37.6 \\
\midrule
\multicolumn{5}{l}{\textit{+ VDrop (mask variants)}} \\
Region & One view  & 30 & 85.2 & {32.6} \\
Region & One view  & 50 & 84.0 & \textbf{40.0} \\
Region & One view  & 80 & \textbf{85.9} & {36.5} \\
Region & Two views & 50 & 83.8 & 36.8 \\
Random & One view  & 50 & 83.1 & 35.6 \\
\midrule
\multicolumn{5}{l}{\textit{+ VDrop, no warmup}} \\
Region & One view  & 30 & 81.9 & 32.2 \\
Region & One view  & 50 & 81.6 & 33.7 \\
\bottomrule[1.5pt]
\end{tabular}%
}
\caption{\textbf{VDrop ablation} on Panoramic visual thinking. \textbf{ID} is the mean over the four COSMIC subtasks; \textbf{OOD} is the out-of-domain mean defined in §\ref{sec:benchmarks}. Bold marks the best per column.}
\label{tab:vdrop_ablation}
\end{table}

\subsection{No-Think Ablation on SenseNova-U1}
\label{sec:appendix_sensenova_nothink}

As a further control for the cross-architecture validation (§\ref{sec:sensenova}), we train a \emph{No-Think} SenseNova-U1 checkpoint on the same spatial data with the thinking-image removed from the target sequence (direct answering), using the same partial-unfreeze recipe as the other SenseNova-U1 variants (§\ref{sec:impl}). Table~\ref{tab:sensenova_nothink} compares it against the untrained model, Panoramic SFT, and Panoramic SFT with VDrop on the OOD benchmarks.

We exclude BLINK from this comparison because the No-Think checkpoint's BLINK score does not measure multi-view reasoning: the checkpoint collapses to a constant predictor, answering ``B'' for every question regardless of its content. The collapse persists under temperature sampling at two seeds, and when we randomly swap the A/B option contents per question, the checkpoint still answers ``B'' on nearly all items and its accuracy falls to chance. This collapse occurs only for the No-Think checkpoint; no visual-thinking checkpoint exhibits it.
On the remaining four benchmarks, VDrop is the strongest setting ($35.5$), above No-Think ($35.0$), the untrained model ($34.3$), and Panoramic SFT ($33.4$).
Moreover, VDrop improves over Panoramic SFT on every individual benchmark.

\subsection{Generate-then-Blind Probe: Setup and Details}
\label{sec:appendix_generate_then_blind}

The generate-then-blind probe in §\ref{sec:generate_then_blind} asks a causal question about the role of the generated thinking-image at inference: does the answer pathway genuinely depend on it, or does it attend primarily to the original input views regardless of what the thinking-image contains?
We compare two models that differ only in their SFT objective:
\textit{(i)} \textbf{Standard SFT}, the visual-thinking SFT pipeline with no VDrop, where the thinking-image tokens contribute to the generation loss so the model is trained both to produce the image and to answer conditioned on it;
\textit{(ii)} \textbf{VDrop SFT}, the same data and recipe with VDrop applied during training, so the answer span must route masked spatial evidence through the generated thinking-image.
Both models start from the same BAGEL initialisation and are fine-tuned on the same 8K Infinigen training set (§\ref{sec:data}).

\paragraph{Intervention procedure.}
For each test item we (a) generate the thinking-image autoregressively as in normal inference, letting its tokens enter the KV cache, and (b) before answer decoding begins, set the attention weights from answer queries to thinking-image positions to $-\infty$ before softmax.
The answer span therefore attends only to $V_1$, $V_2$, and the question, so the model still ``thinks'' by producing the image but cannot read it back when answering.

\paragraph{Evaluation.}
We compare each blinded run against the same model's unblinded baseline on four OOD benchmarks (Figure~\ref{fig:generate_then_blind}).
In-distribution accuracy on COSMIC is largely saturated by visual-thinking SFT and is therefore an insensitive test of causal dependence; the OOD benchmarks, where models still have substantial headroom, provide a cleaner signal for whether the thinking-image is genuinely load-bearing.

\begin{figure}
    \centering
    \includegraphics[width=\linewidth]{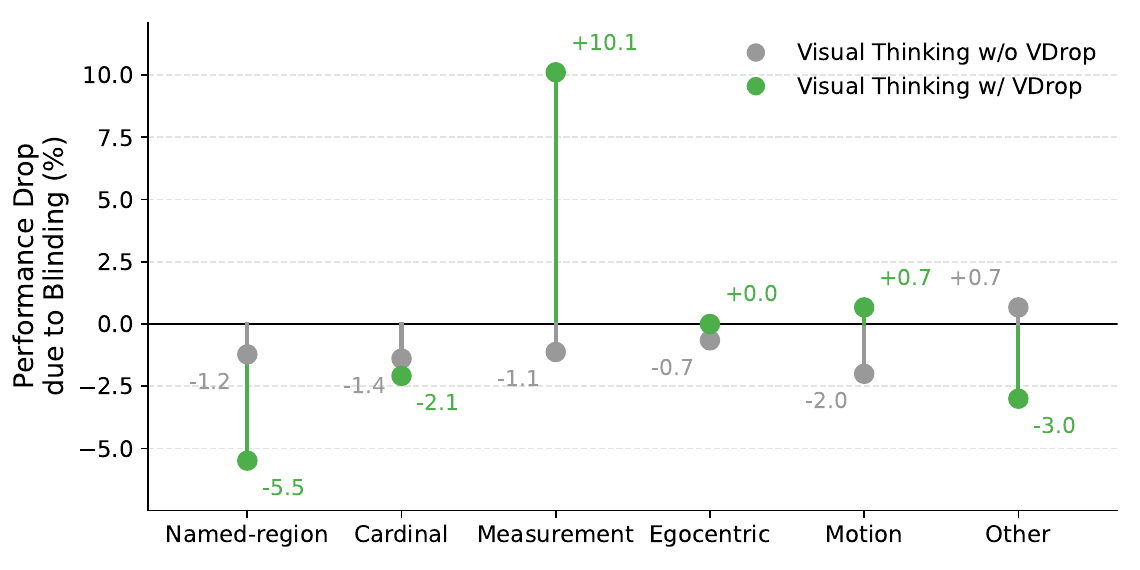}
    \caption{\textbf{Generate-then-blind probe on MMSI, by question evidence category.} Accuracy drop when the generated thinking-image is blinded at answer time; a larger positive value means more dependence on the thinking-image. The VDrop-trained model shows a large drop only on Measurement, whose questions are answered by visually aligning the two input views, while standard SFT is unaffected throughout.}
    \label{fig:generate_then_blind_mmsi}
\end{figure}

\paragraph{MMSI breakdown.}
MMSI's questions ask for spatially heterogeneous evidence, so a single overall accuracy obscures whether the generated thinking-image helps. We re-partition the $1{,}000$ MMSI questions into six disjoint categories by the kind of evidence each requires. \emph{Measurement} questions need visual comparison (``which is wider'', ``how many''), exactly the operation enabled by aligning $V_1$ and $V_2$ in one panorama. \emph{Egocentric} questions ask for viewer-relative direction, which a wider-field-of-view panorama makes visible. \emph{Named-region} and \emph{Cardinal} questions depend on symbolic information a panorama cannot encode: room labels (``kitchen'') and absolute compass frames (``north of the desk''). \emph{Motion} questions concern camera or object movement, and \emph{Other} covers the remainder, mostly non-spatial counting. Each question is assigned to the first matching category in the order Motion, Named-region, Cardinal, Measurement, Egocentric, Other.

Figure~\ref{fig:generate_then_blind_mmsi} reports the blinding-induced accuracy drop per category. Standard SFT is unaffected in every category (within $\pm 2$ pp of zero, max $|z| = 0.6$\footnote{We test each category's blinding effect against zero with a paired test over per-question accuracy deltas; $z$ is the effect divided by its standard error, and $|z| \gtrsim 2$ corresponds to $p < 0.05$.}), confirming the thinking-image is inert without VDrop. The VDrop-trained model shows a large, significant drop on Measurement ($+10.1$ pp, $z = 2.24$, $p = 0.025$), the category answered by visually aligning the two views; on Named-region and Cardinal, whose answers a panorama cannot encode, blinding has no positive effect. 
VDrop thus teaches the model to genuinely use the thinking-image, and this reliance surfaces precisely on questions that call for visual reasoning, while standard training leaves it largely ignored.

\subsection{Answer-Token Attention Probe: Setup and Details}
\label{sec:appendix_attention_probe}

The attention probe in §\ref{sec:vdrop_attention} asks where the answer span's visual attention is directed during answer generation, and whether VDrop measurably increases the share directed to the model's own generated thinking-image.

\paragraph{Models compared.}
We extract attention weights from the answer-generation step across all decoder layers and compare \textbf{Standard SFT} (visual-thinking SFT without VDrop) and \textbf{VDrop SFT} (visual-thinking SFT with VDrop), both trained on the same data. We additionally include \textbf{vanilla BAGEL} as a reference. Vanilla BAGEL is not trained for visual thinking and does not reliably emit a thinking-image before answering; to obtain a comparable thinking-image span, we force generation by injecting the image-generation token into the decoding stream, after which it produces a thinking-image and then an answer. This makes the thinking-image span well-defined for all three models, so the share metric below is computed identically across them. We note that vanilla BAGEL's forced thinking-image is not optimised for the task and serves only as an untrained reference point.

\paragraph{Probe metric.}
To focus on visual evidence specifically, we normalise attention over the named visual spans only: the two input views $V_1$ and $V_2$, plus the generated thinking-image visual tokens $\textsc{vt\_all}$.
For each decoder layer $\ell$ and each evaluation example, the quantity of interest is the per-layer \emph{thinking-image share among visual evidence}:
\begin{equation*}
    \rho_{\text{vt},\ell} \;=\; \frac{\mathrm{attn}_\ell(\textsc{vt\_all})}{\mathrm{attn}_\ell(V_1) + \mathrm{attn}_\ell(V_2) + \mathrm{attn}_\ell(\textsc{vt\_all})},
\end{equation*}
where $\mathrm{attn}_\ell(\cdot)$ is the answer-token attention mass on the named visual span at layer $\ell$, averaged across all answer-token positions and attention heads.
By construction $\rho_{\text{vt},\ell} \in [0, 1]$, with higher values indicating that the answer query relies more on the generated thinking-image than on the input views at layer $\ell$.
We report the mean of $\rho_{\text{vt},\ell}$ across all evaluation examples.

\begin{figure}[t]
    \centering
    \includegraphics[width=\linewidth]{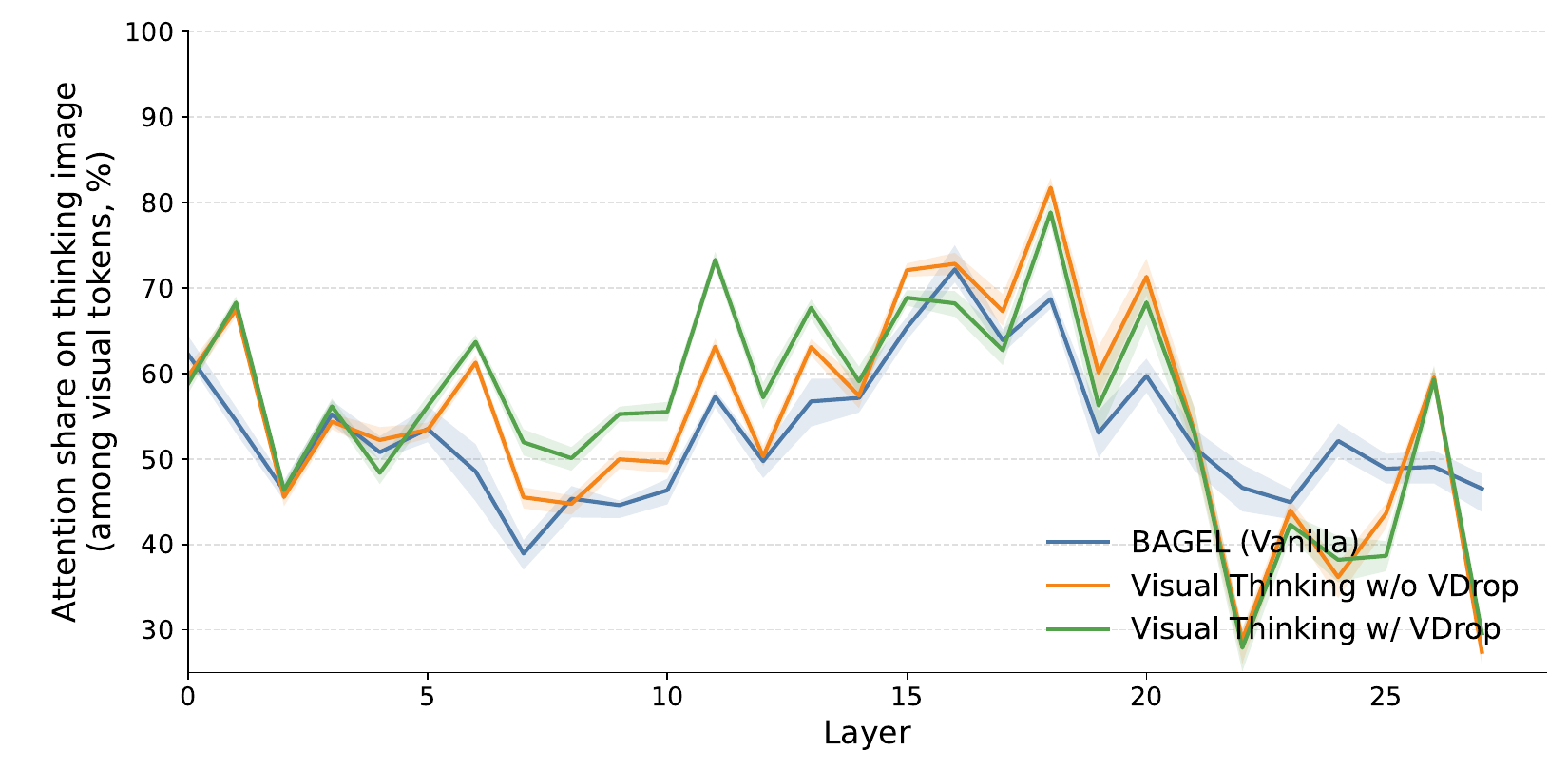}
    \caption{\textbf{Mean answer-token attention on thinking-image tokens across decoder layers (STARE).} The VDrop-trained model places more attention on the generated thinking-image than the standard SFT model, especially in early and mid layers, indicating that VDrop shifts the answer pathway toward the thinking-image.}
    \label{fig:vt_share_by_layer_stare}
\end{figure}

\paragraph{STARE-Perspective: per-layer attention share.}
Figure~\ref{fig:vt_share_by_layer_stare} reproduces the $\rho_{\text{vt},\ell}$ measurement on 250 STARE-Perspective examples. The pattern matches BLINK in the early and middle layers: averaged over the first 14 layers of the decoder, VDrop places $+3.5$ pp more attention on the thinking-image than standard SFT. Over all layers the gain is smaller ($+0.9$ pp over standard SFT), as the effect is concentrated early, with the two models converging in the late decoder layers. VDrop's increased engagement with the thinking-image is therefore localised to the early-to-middle layers rather than spread uniformly across the network, and this localisation is consistent across BLINK and STARE.

\subsection{Theoretical Grounding of the L--I Framework}
\label{sec:appendix_li_grounding}

{This section expands the imitation-learning view of the Learnability--Informativeness framework introduced in §\ref{sec:l_i_tradeoff}.}

\paragraph{Visual-thinking SFT is behavioural cloning of the target intermediate.}
SFT trains the UMM to reproduce a \emph{target} thinking-image from the inputs and then answer conditioned on it, i.e., it clones a demonstrated intermediate rather than discovering one, exactly as instruction tuning clones demonstrated responses~\cite{stiennon2020learning,ouyang2022training}.
Imitation-learning theory bounds the value of a cloned policy by two \emph{independent} error terms.
First, cloning cannot exceed the demonstrator it imitates: the learned policy's value is bounded by the value of the imitated target, up to terms that vanish as imitation becomes exact~\cite{rajaraman2020toward}.
Second, when the target is not realizable by the learner's model class, an irreducible misspecification error remains no matter how much data is used~\cite{espinosa2025efficient}.

\paragraph{The two bounds are the two axes.}
\textbf{Informativeness} $I(T)$ instantiates the first term: it measures the value of a \emph{perfect} instance of type $T$ for the downstream answer, which upper-bounds what any model cloning $T$ can gain. We isolate it by feeding \emph{ground-truth} renders to a frozen VLM answerer (Table~\ref{tab:two_reader_oracle}), removing the UMM's generation quality from the measurement entirely.
\textbf{Learnability} $L(T)$ instantiates the second term: it measures how much of that target the UMM actually realises. We isolate it by feeding the UMM's \emph{generated} images to the same frozen answerer (Table~\ref{tab:generation_understanding}); any drop from the ground-truth to the generated measurement is realisation error, not a deficiency of the representation itself.

\paragraph{The framework is falsifiable and practical.}
The imitation-learning view yields a testable prediction: a thinking-image type helps end-to-end only if both terms are favourable, and the two probes, which are cheap to run before any end-to-end training, should predict the end-to-end ranking. Our results confirm this: top-down is informative but only partially learnable, point matching is learnable but uninformative, and only panorama scores on both axes, matching the end-to-end ordering in Table~\ref{tab:vdrop_with_without}. When a strategy underperforms, the probes also pinpoint which axis is the bottleneck, indicating whether to improve the target representation or the model's ability to generate it.

\subsection{Qualitative Analysis}
\label{sec:appendix_qualitative}
\begin{figure*}[t]
\centering
\resizebox{\textwidth}{!}{%
\begin{tikzpicture}[x=1cm, y=1cm, >=Latex, font=\sffamily\small]

\definecolor{myviolet}{RGB}{135,95,180}

\def\colQ    { 0.0}
\def\qWidth  {3.8cm}
\def\colInOne{ 5.8}\def\colInTwo{ 9.4}
\def\colPano {13.1}\def\colPM  {16.7}\def\colCV {19.9}
\def\rowHdr  {10.5}
\def\rowOne  { 9.0}\def\rowTwo { 6.6}\def\rowThree{ 4.2}\def\rowFour{ 1.8}
\def\rowOneL { 7.95}\def\rowTwoL{ 5.55}\def\rowThreeL{ 3.15}\def\rowFourL{ 0.75}
\def\imgH{1.55cm}
\def\wPano{3.4cm}
\def\wPM{3.4cm}

\node[anchor=west, text=black!70, font=\bfseries\footnotesize] at (\colQ,\rowHdr) {Question \& options};
\node[text=myviolet,font=\bfseries\footnotesize] at (\colInOne,\rowHdr) {Input View 1};
\node[text=myviolet,font=\bfseries\footnotesize] at (\colInTwo,\rowHdr) {Input View 2};
\node[text=black!65,font=\bfseries\footnotesize] at (\colPano ,\rowHdr) {Panoramic};
\node[text=black!65,font=\bfseries\footnotesize] at (\colPM   ,\rowHdr) {Point Matching};
\node[text=black!65,font=\bfseries\footnotesize] at (\colCV   ,\rowHdr) {Top-down View};

\draw[dashed,black!30,line width=0.4pt] (11.0,0.0) -- (11.0,10.0);

\node[anchor=west, text width=\qWidth, align=left, font=\scriptsize] at (\colQ,\rowOne)
    {\textbf{Anchor.} Which of the following objects is visible from both views of the room?\\
     A) cabinet near a brown bed\\
     B) cabinet near a white desk\\
     C) lamp next to a gray plant\\
     \textcolor{checkgreen}{D) cabinet next to a purple lamp \cmark}};
\node[draw=black!35,rounded corners=2pt,inner sep=0pt] at (\colInOne,\rowOne) {\includegraphics[height=\imgH]{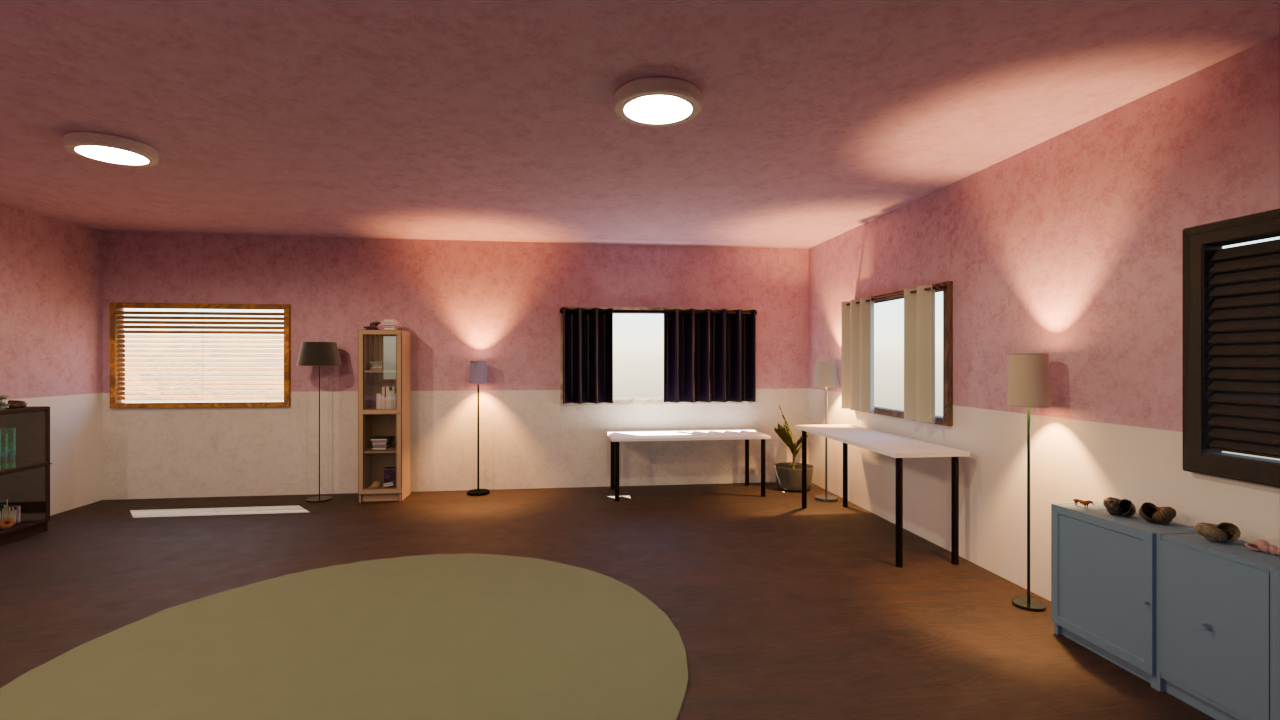}};
\node[draw=black!35,rounded corners=2pt,inner sep=0pt] at (\colInTwo,\rowOne) {\includegraphics[height=\imgH]{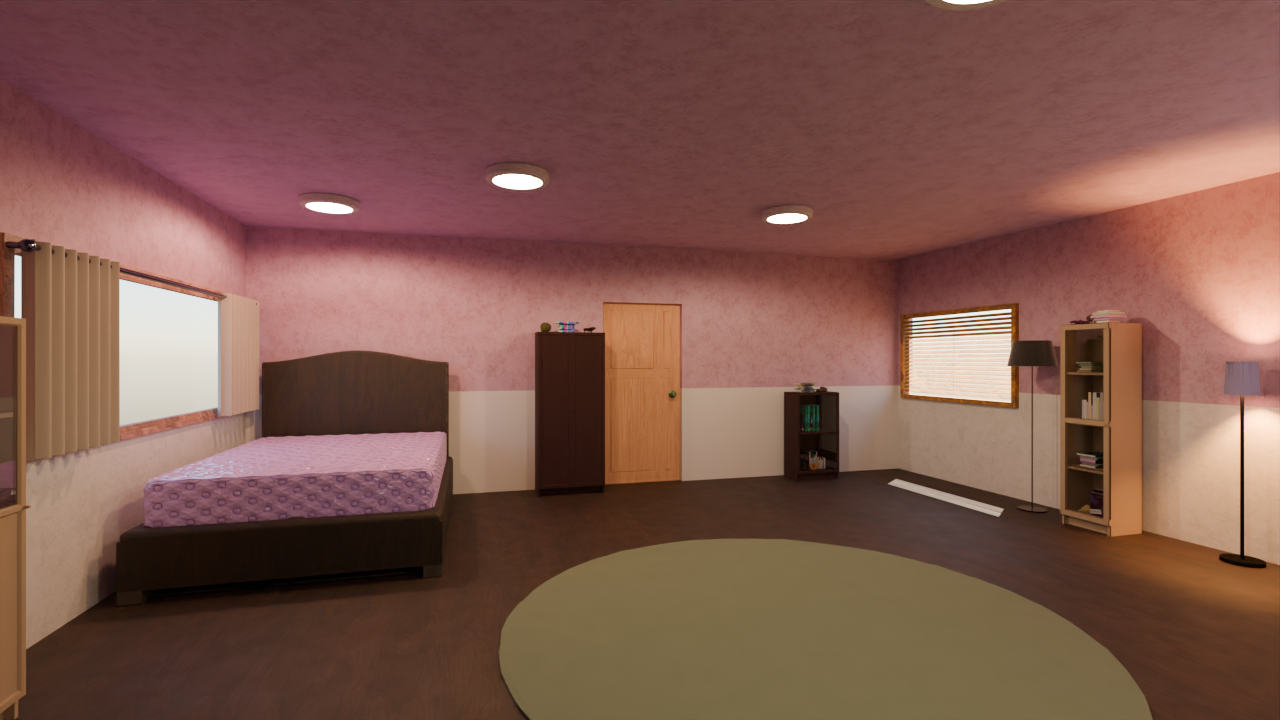}};
\node[draw=black!35,rounded corners=2pt,inner sep=0pt] at (\colPano,\rowOne) {\includegraphics[height=\imgH,width=\wPano]{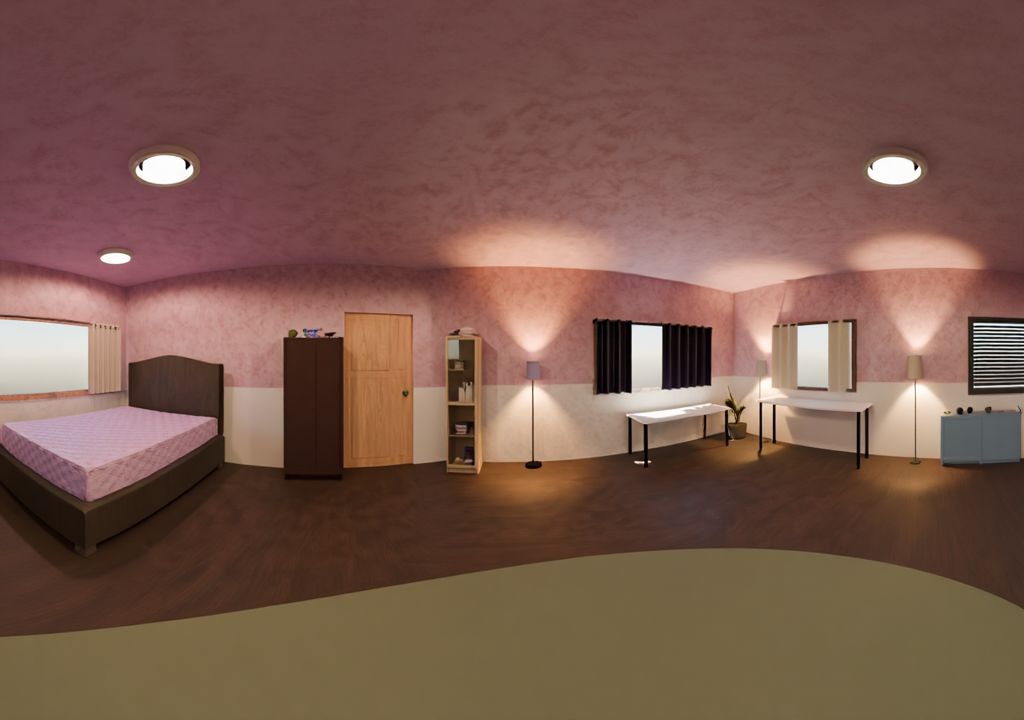}};
\node[font=\scriptsize] at (\colPano,\rowOneL) {pred: D \cmark};
\node[draw=black!35,rounded corners=2pt,inner sep=0pt] at (\colPM,\rowOne) {\includegraphics[height=\imgH,width=\wPM]{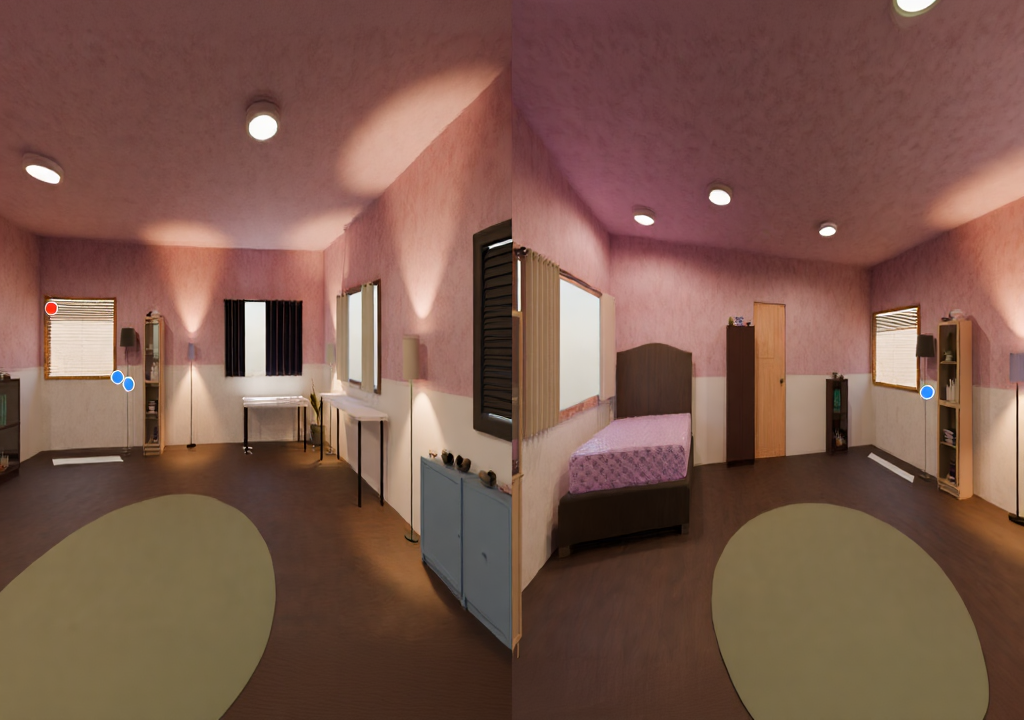}};
\node[font=\scriptsize] at (\colPM,\rowOneL) {pred: D \cmark};
\node[draw=black!35,rounded corners=2pt,inner sep=0pt] at (\colCV,\rowOne) {\includegraphics[height=\imgH]{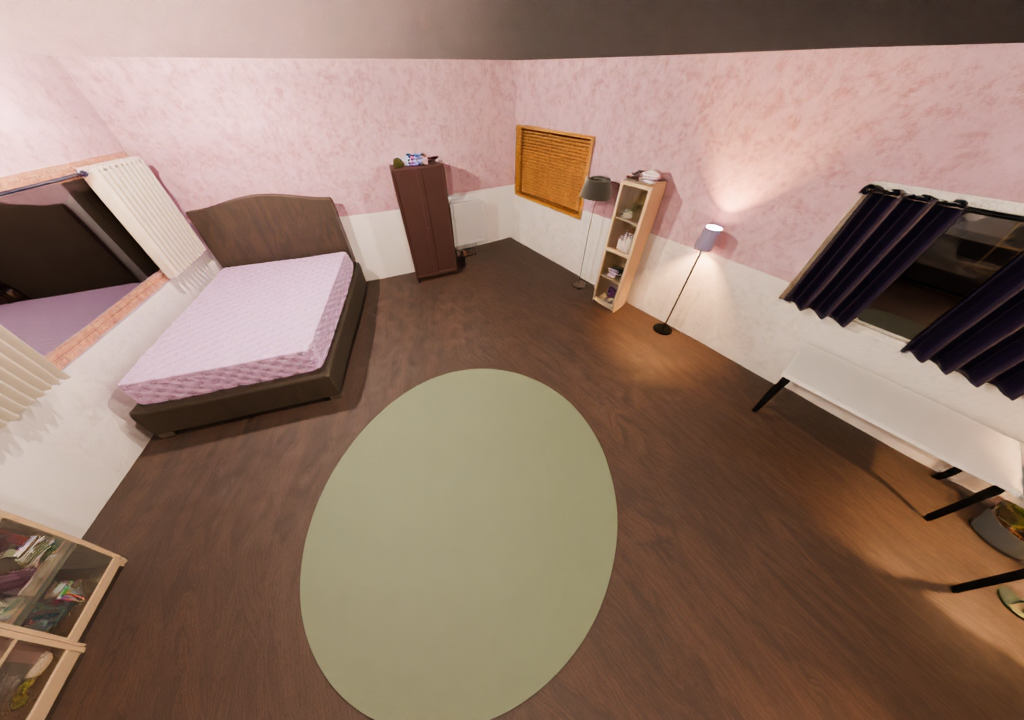}};
\node[font=\scriptsize] at (\colCV,\rowOneL) {pred: D \cmark};

\node[anchor=west, text width=\qWidth, align=left, font=\scriptsize] at (\colQ,\rowTwo)
    {\textbf{Counting.} What is the total number of cabinets in the room?\\
     \textcolor{checkgreen}{A) 5 \cmark}\\
     B) 6\\
     C) 1\\
     D) 7};
\node[draw=black!35,rounded corners=2pt,inner sep=0pt] at (\colInOne,\rowTwo) {\includegraphics[height=\imgH]{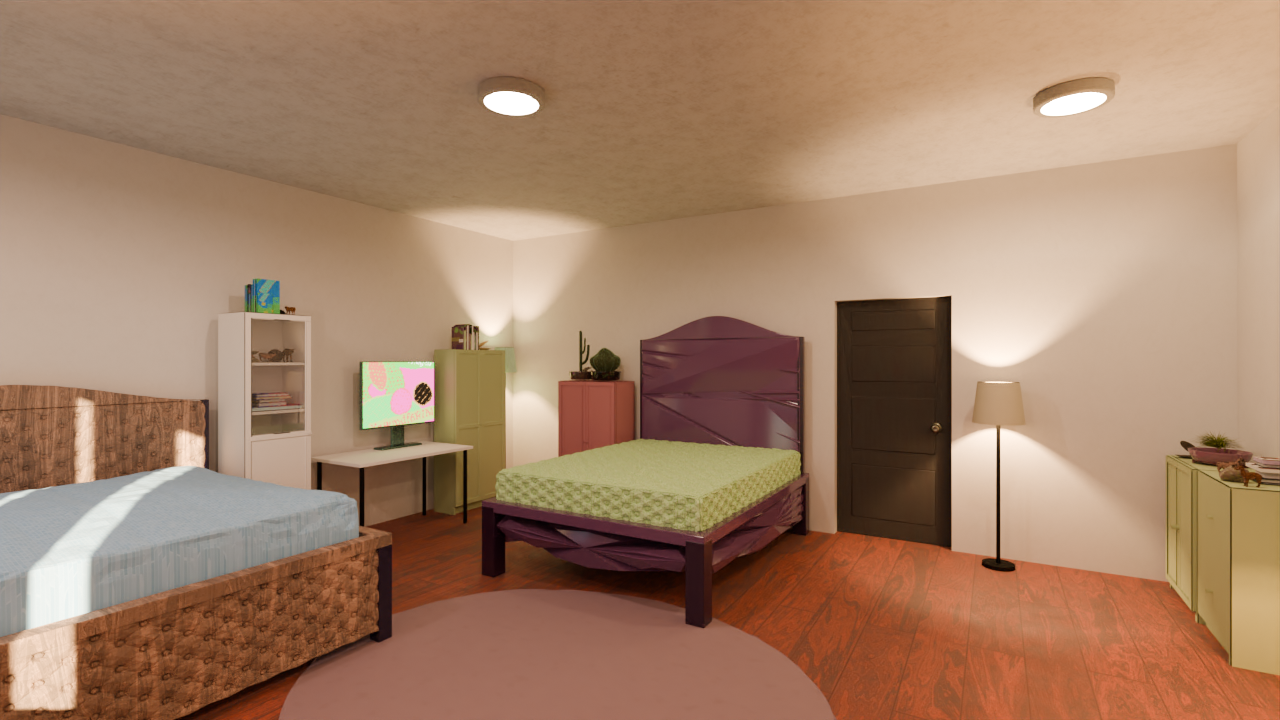}};
\node[draw=black!35,rounded corners=2pt,inner sep=0pt] at (\colInTwo,\rowTwo) {\includegraphics[height=\imgH]{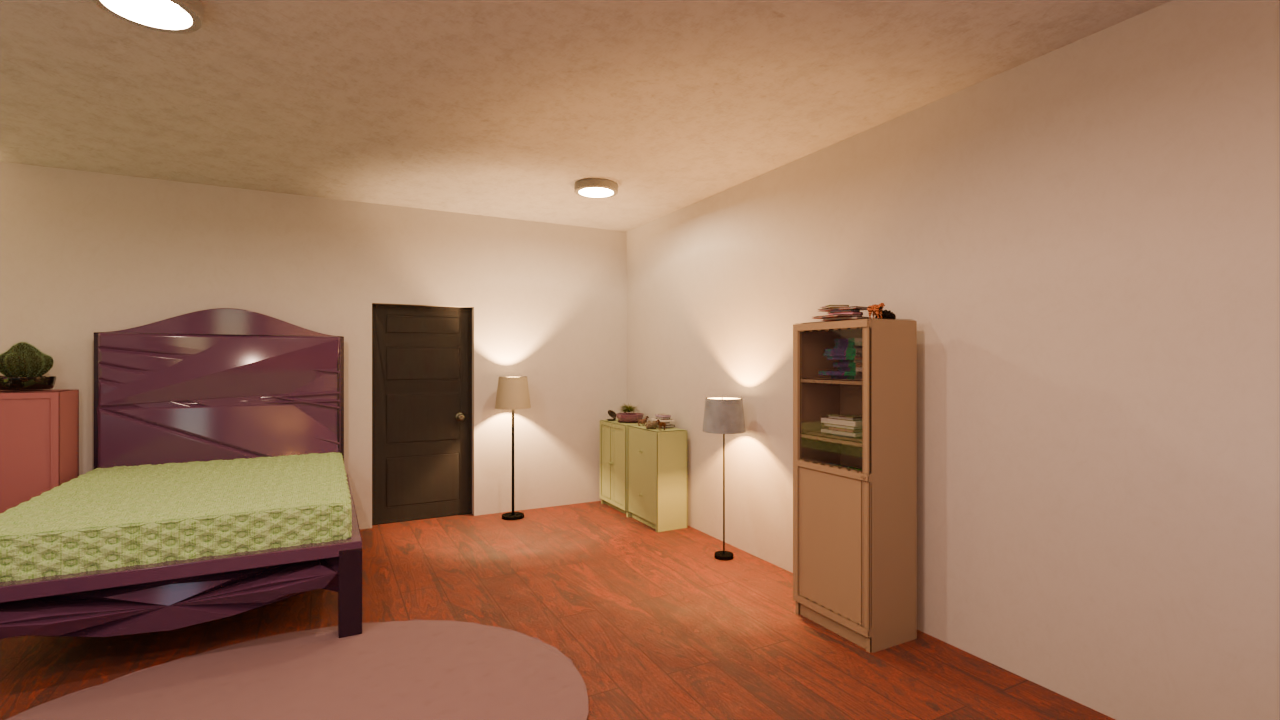}};
\node[draw=black!35,rounded corners=2pt,inner sep=0pt] at (\colPano,\rowTwo) {\includegraphics[height=\imgH,width=\wPano]{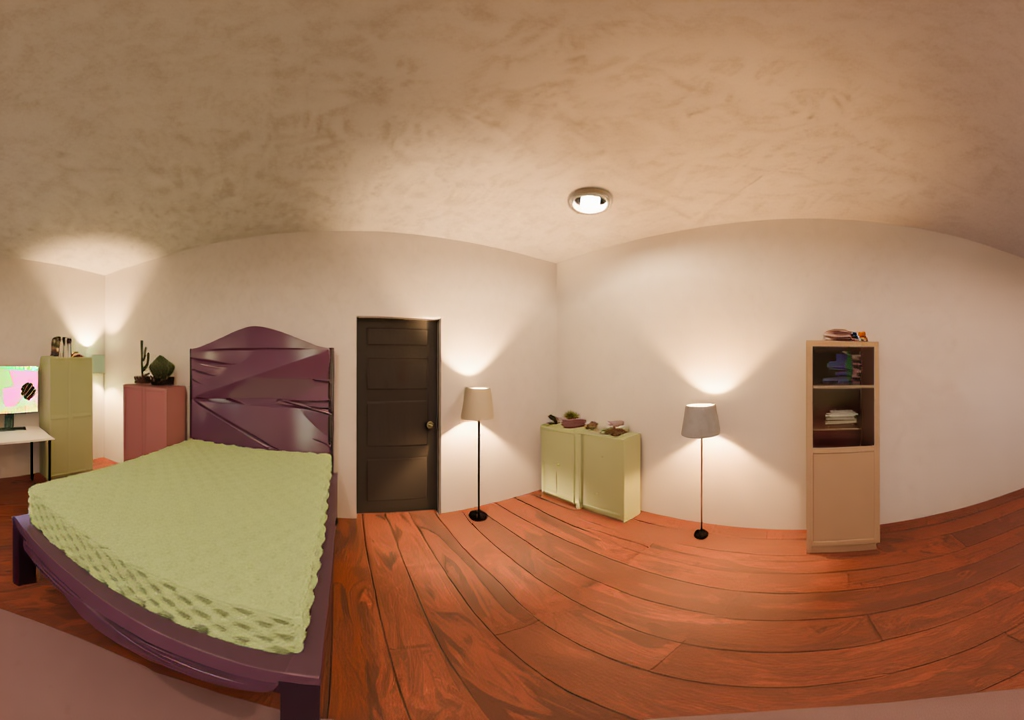}};
\node[font=\scriptsize] at (\colPano,\rowTwoL) {pred: A \cmark};
\node[draw=black!35,rounded corners=2pt,inner sep=0pt] at (\colPM,\rowTwo) {\includegraphics[height=\imgH,width=\wPM]{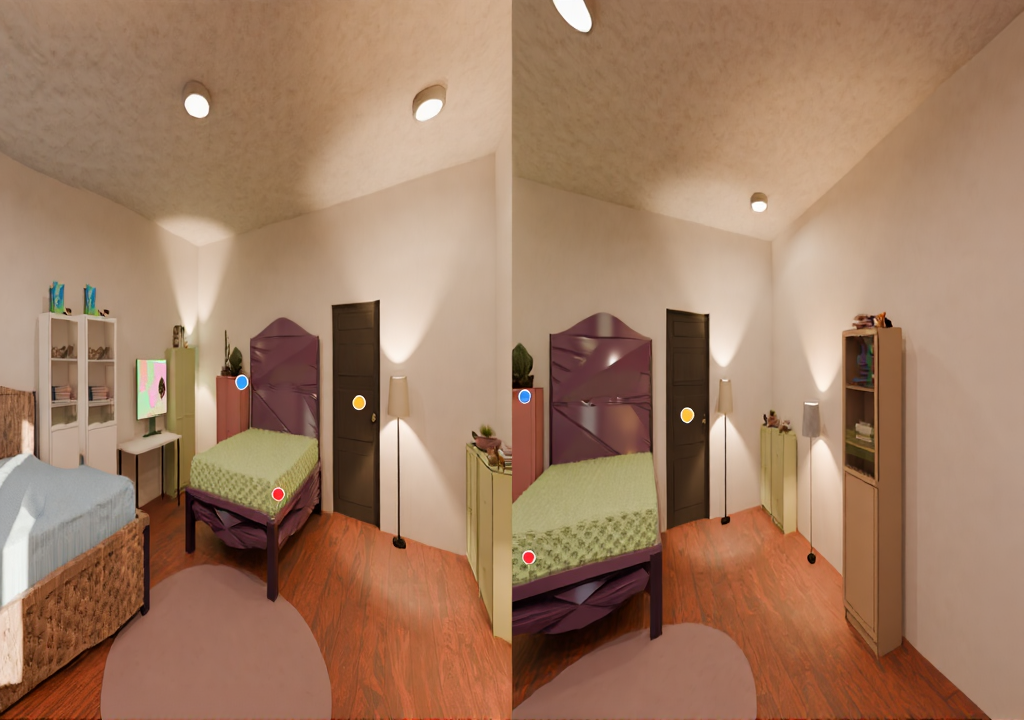}};
\node[font=\scriptsize] at (\colPM,\rowTwoL) {pred: A \cmark};
\node[draw=black!35,rounded corners=2pt,inner sep=0pt] at (\colCV,\rowTwo) {\includegraphics[height=\imgH]{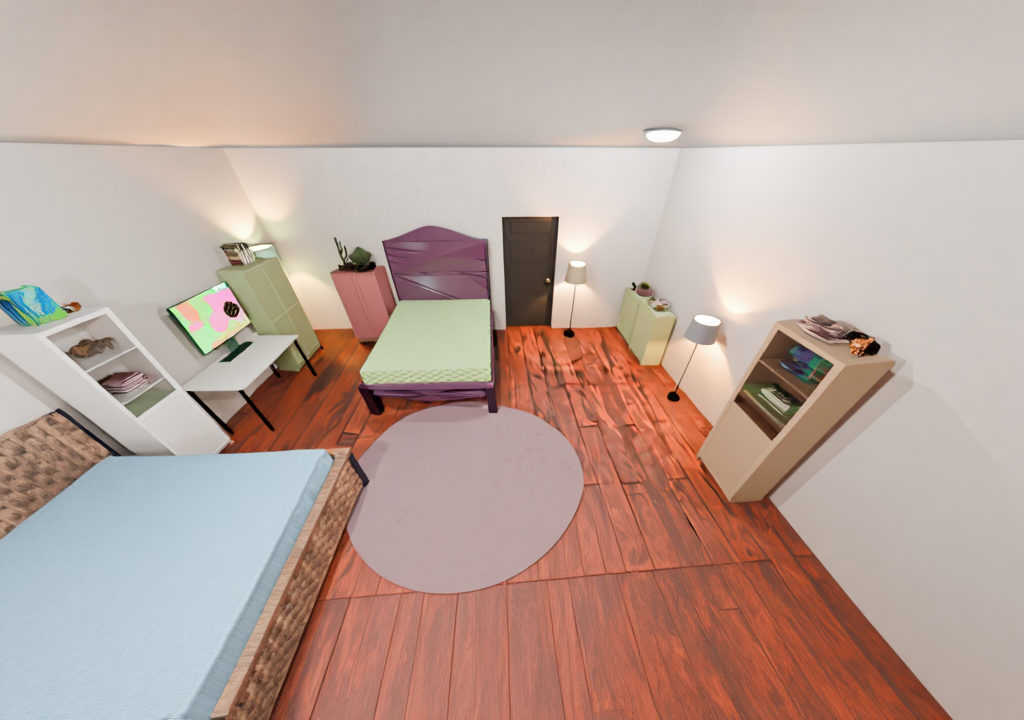}};
\node[font=\scriptsize] at (\colCV,\rowTwoL) {pred: A \cmark};

\node[anchor=west, text width=\qWidth, align=left, font=\scriptsize] at (\colQ,\rowThree)
    {\textbf{Relative Distance.} Which of the following objects is the farthest from the dark brown cabinet?\\
     A) blue cabinet near a purple bed\\
     \textcolor{checkgreen}{B) green cabinet near a black desk \cmark}\\
     C) lamp next to a blue cabinet\\
     D) brown door next to a beige lamp};
\node[draw=black!35,rounded corners=2pt,inner sep=0pt] at (\colInOne,\rowThree) {\includegraphics[height=\imgH]{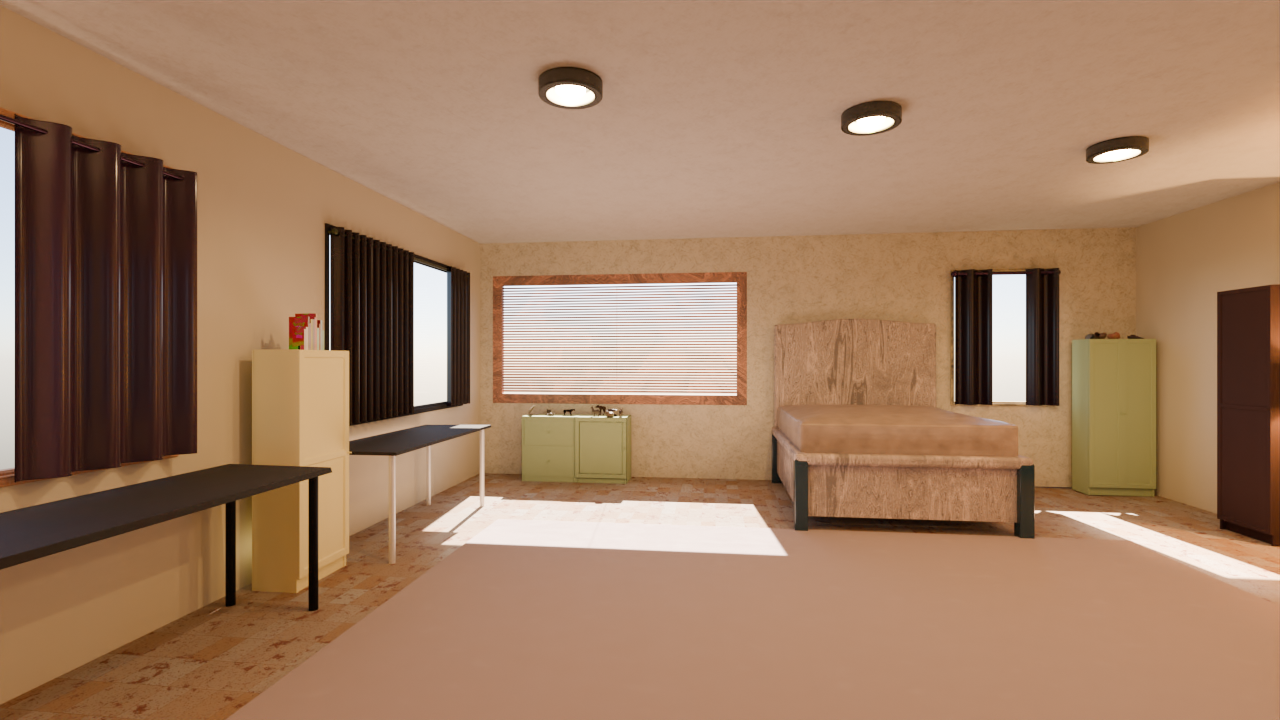}};
\node[draw=black!35,rounded corners=2pt,inner sep=0pt] at (\colInTwo,\rowThree) {\includegraphics[height=\imgH]{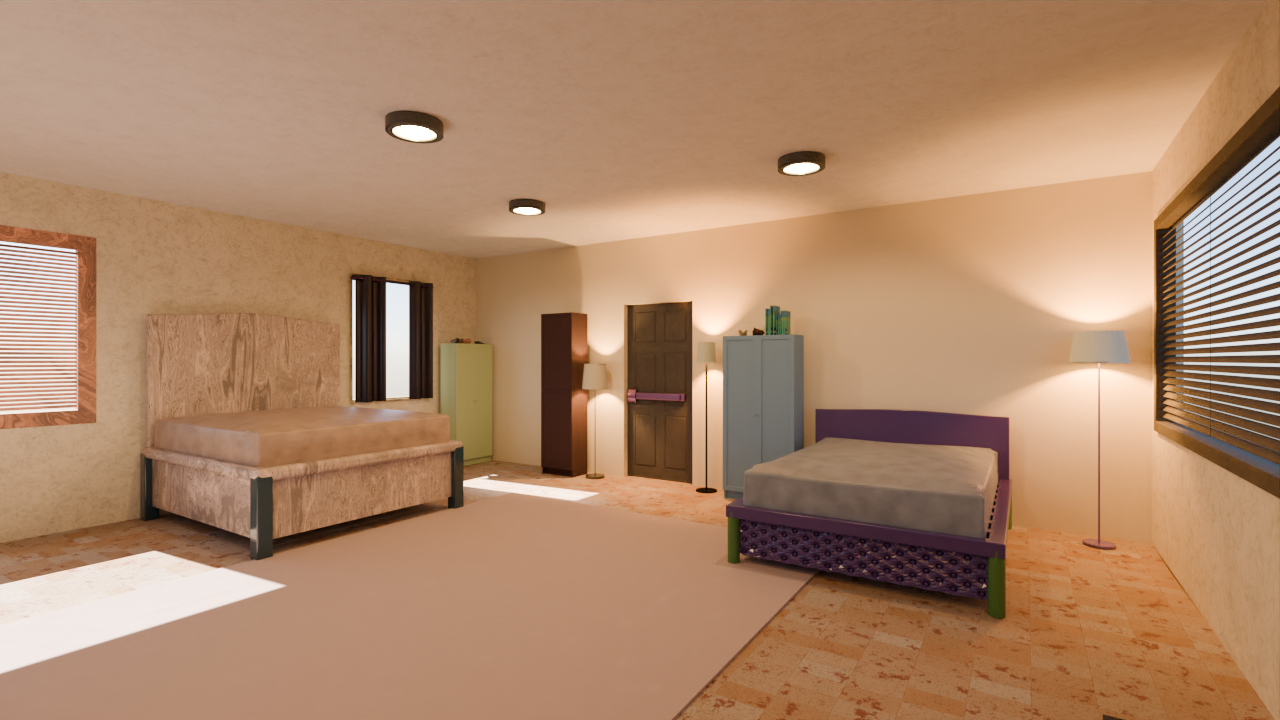}};
\node[draw=black!35,rounded corners=2pt,inner sep=0pt] at (\colPano,\rowThree) {\includegraphics[height=\imgH,width=\wPano]{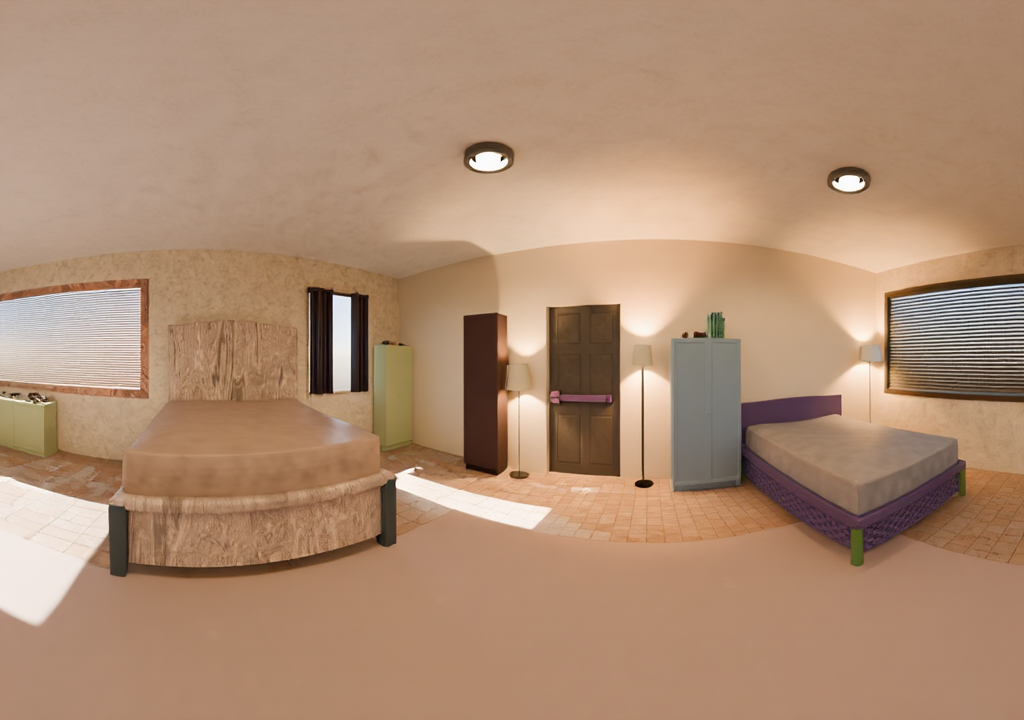}};
\node[font=\scriptsize] at (\colPano,\rowThreeL) {pred: B \cmark};
\node[draw=black!35,rounded corners=2pt,inner sep=0pt] at (\colPM,\rowThree) {\includegraphics[height=\imgH,width=\wPM]{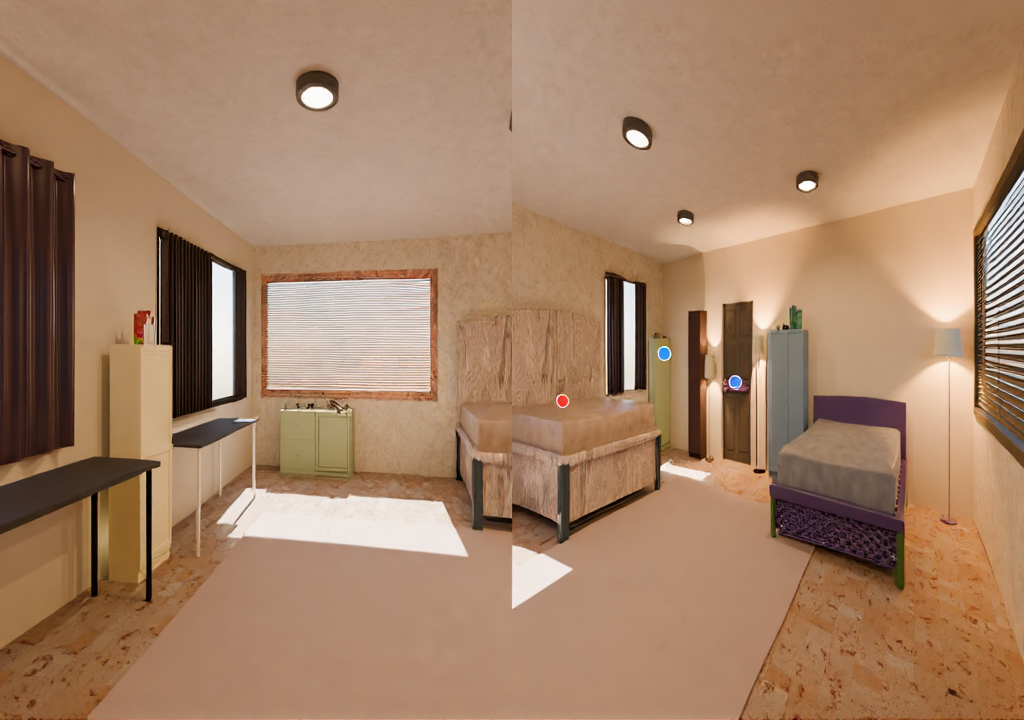}};
\node[font=\scriptsize] at (\colPM,\rowThreeL) {pred: C \xmark};
\node[draw=black!35,rounded corners=2pt,inner sep=0pt] at (\colCV,\rowThree) {\includegraphics[height=\imgH]{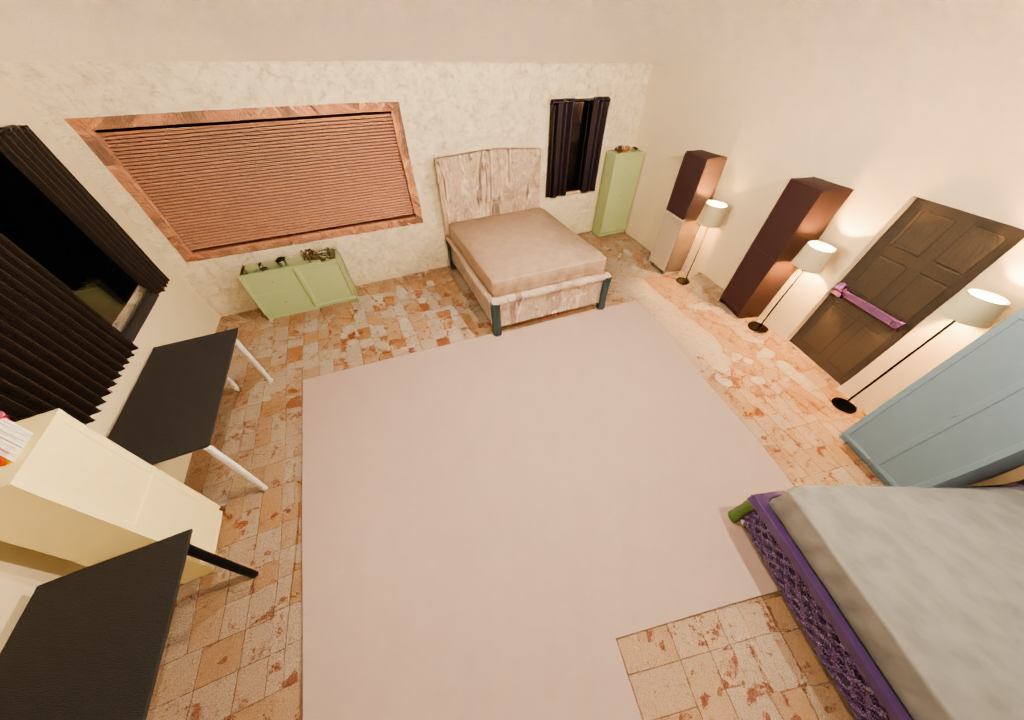}};
\node[font=\scriptsize] at (\colCV,\rowThreeL) {pred: C \xmark};

\node[anchor=west, text width=\qWidth, align=left, font=\scriptsize] at (\colQ,\rowFour)
    {\textbf{Relative Direction.} Relative to you, in which direction is the white desk from the viewpoint of the second image?\\
     A) right\\
     B) front\\
     \textcolor{checkgreen}{C) left \cmark}\\
     D) behind};
\node[draw=black!35,rounded corners=2pt,inner sep=0pt] at (\colInOne,\rowFour) {\includegraphics[height=\imgH]{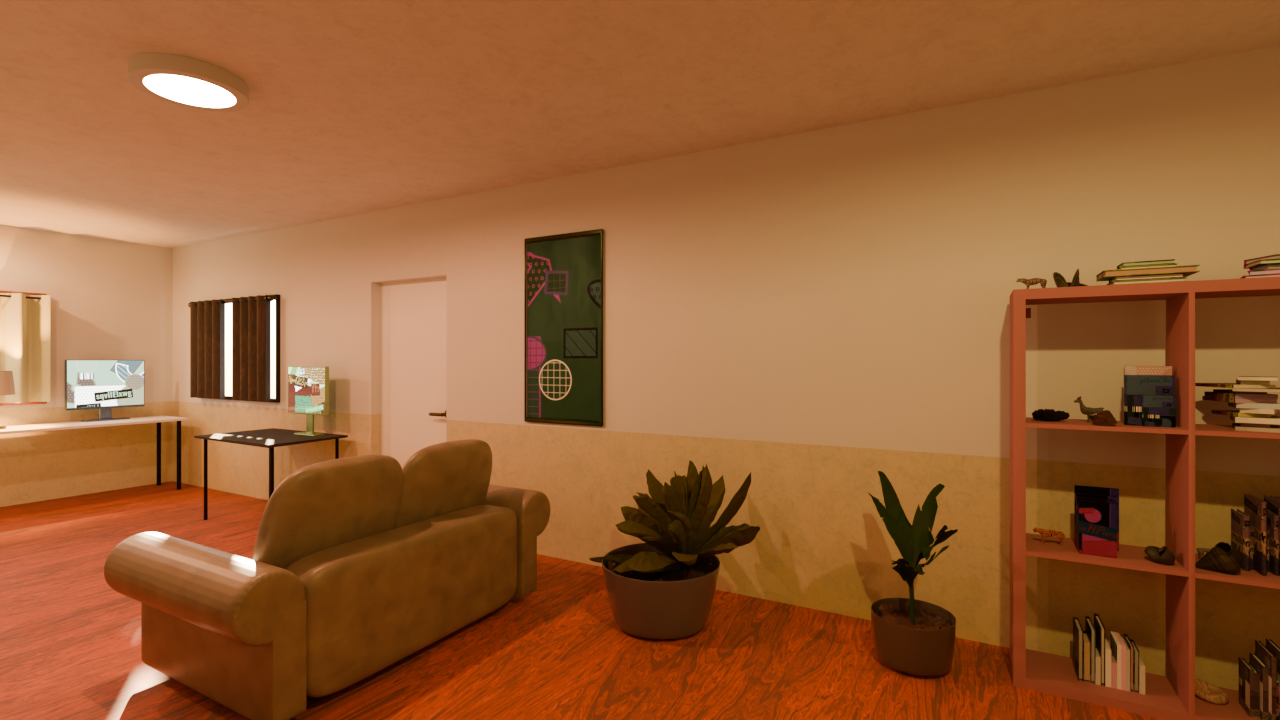}};
\node[draw=black!35,rounded corners=2pt,inner sep=0pt] at (\colInTwo,\rowFour) {\includegraphics[height=\imgH]{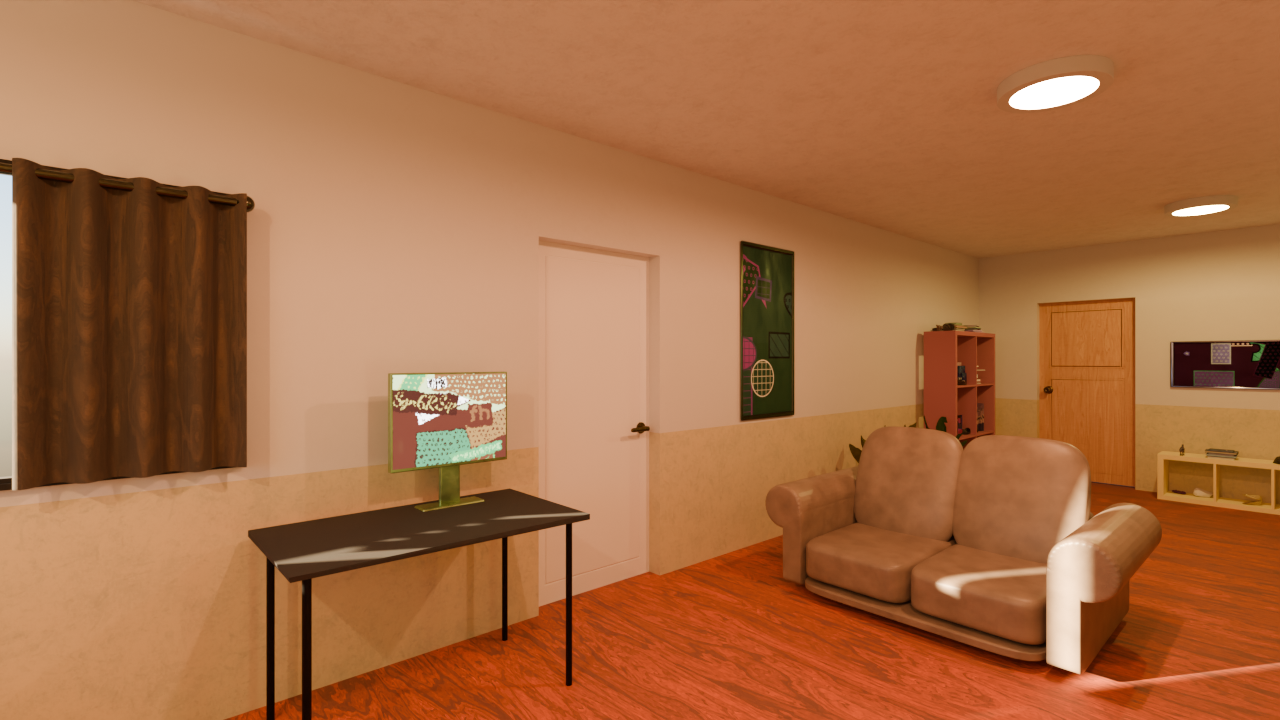}};
\node[draw=black!35,rounded corners=2pt,inner sep=0pt] at (\colPano,\rowFour) {\includegraphics[height=\imgH,width=\wPano]{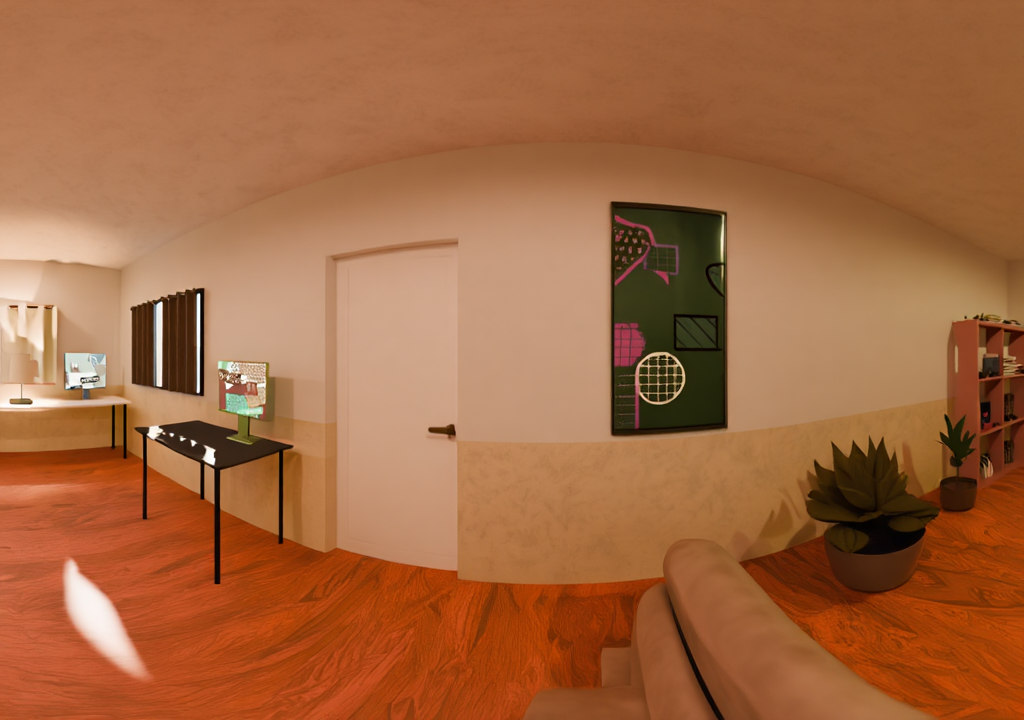}};
\node[font=\scriptsize] at (\colPano,\rowFourL) {pred: C \cmark};
\node[draw=black!35,rounded corners=2pt,inner sep=0pt] at (\colPM,\rowFour) {\includegraphics[height=\imgH,width=\wPM]{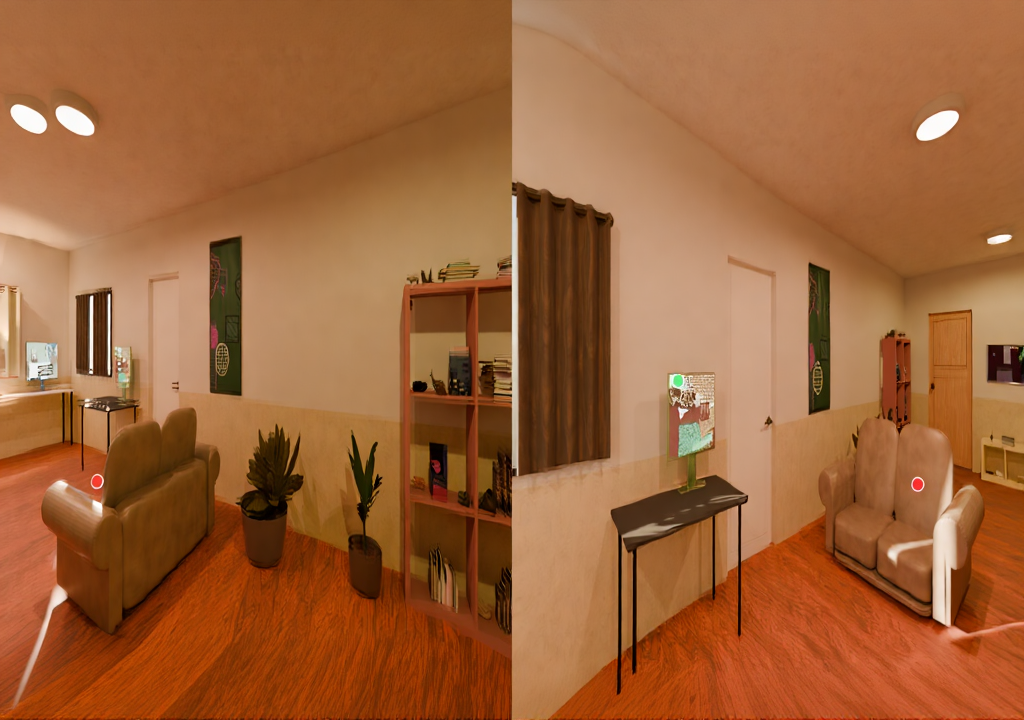}};
\node[font=\scriptsize] at (\colPM,\rowFourL) {pred: A \xmark};
\node[draw=black!35,rounded corners=2pt,inner sep=0pt] at (\colCV,\rowFour) {\includegraphics[height=\imgH]{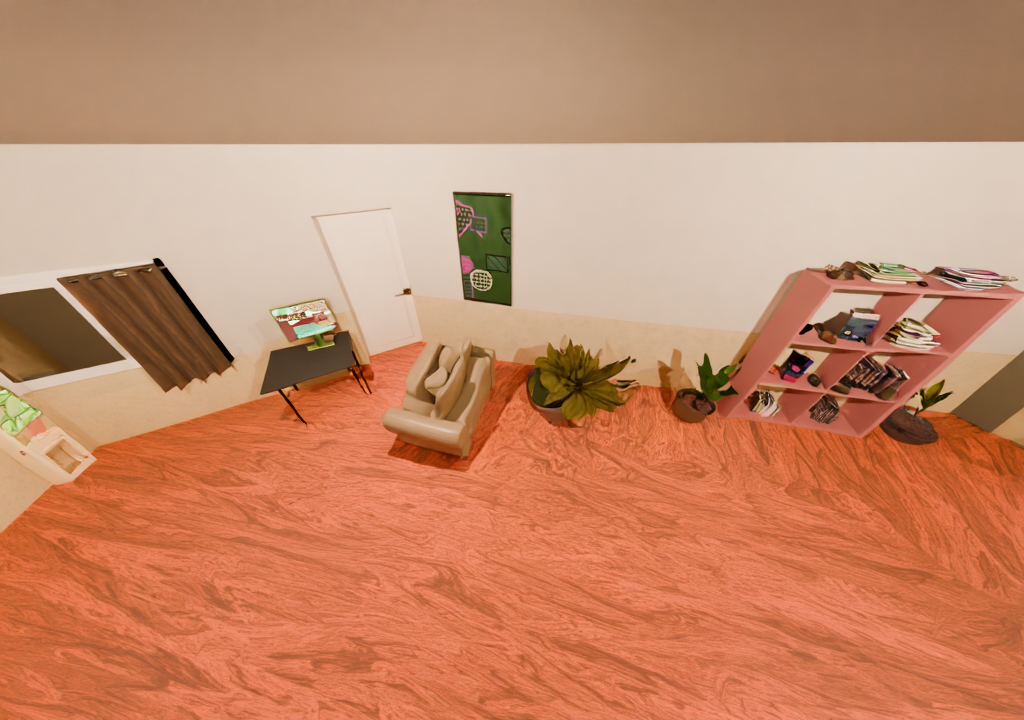}};
\node[font=\scriptsize] at (\colCV,\rowFourL) {pred: A \xmark};

\end{tikzpicture}%
}
\caption{\textbf{Qualitative examples of visual thinking across strategies.}
Four samples, one per subtask (Anchor, Counting, Relative Distance, Relative Direction). Each row shows the question and four options (gold option in \textcolor{checkgreen}{green}), followed by the two input camera views and the generated thinking-image under each strategy (Panoramic, Point Matching, Top-down View). The predicted answer letter and correctness (\cmark{} / \xmark{}) are shown below each thinking-image.}
\label{fig:qualitative_analysis}
\end{figure*}

Figure~\ref{fig:qualitative_analysis} shows the thinking-image our VDrop-trained model generates under each of the three strategies (Panoramic, Point Matching, Top-down) on one example per subtask, together with the question and four options (gold option in \textcolor{checkgreen}{green}). Below each thinking-image we mark the predicted answer and whether it matches the gold label.
The examples illustrate both axes of the L--I analysis (§\ref{sec:l_i_tradeoff}). 
On \emph{informativeness}: panoramic and top-down thinking-images render the scene from a genuinely new viewpoint, directly revealing the joint spatial layout that relative-distance and relative-direction questions require. Point matching instead keeps both input views in their original perspective and overlays correspondence markers; it still helps, by linking objects across the two views, but it conveys cross-view geometry only \emph{indirectly} rather than exposing the full scene directly.
On \emph{learnability}: the point-matching examples also reveal a failure specific to this strategy. The correspondence markers are only a few pixels wide, far smaller than the large-area object relationships that panoramic and top-down images convey, and are correspondingly harder to generate reliably. In the Anchor and Relative-Distance examples, the model places a marker on an object in one view but omits the matching marker in the other, breaking the correspondence the strategy depends on.
Together this matches the aggregate trend in §\ref{sec:vdrop_with_without}, where panoramic visual thinking yields the largest out-of-domain gains.

\graphicspath{{Figures/}}

\end{document}